\documentclass[journal]{IEEEtran}

\ifCLASSINFOpdf
\else
\fi

\usepackage{times}
\usepackage{epsfig}
\usepackage{graphicx}
\usepackage{amsmath}
\usepackage{amssymb}
\usepackage{dsfont}
\usepackage{multirow}
\usepackage{epstopdf}
\usepackage{bmpsize}
\usepackage{enumerate}
\usepackage{hyperref}
\usepackage{array}
\usepackage{color}
\usepackage{multirow}
\usepackage{amsmath}
\usepackage{cite}
\usepackage[misc]{ifsym}
\usepackage{subfig}
\usepackage{authblk}
\usepackage{makecell}
\usepackage{amsthm}

\usepackage{tabularx}
\usepackage{booktabs}
\usepackage{tikz}

\begin{document}
%
\title{Cross-camera Trajectories Help Person Retrieval in a Camera Network}
%
%
%

\author{Xin Zhang,
        Xiaohua~Xie,~\IEEEmembership{Member,~IEEE.} %
        Jianhuang~Lai,~\IEEEmembership{Senior~Member,~IEEE}
        Wei-Shi~Zheng,~\IEEEmembership{Member,~IEEE}
\thanks{Manuscript received XXX XX, XXXX; revised XXX XX, XXXX. This work was supported in part by the National Natural Science Foundation of China (U22A2095, 62072482, 61573387), Key-Area Research and Development Program of Guangdong Province (2019B010155003), and Key-Area Research and Development Program of Guangzhou (202206030003). (Corresponding author: Xiaohua Xie.)}
\thanks{The authors are with the School of Computer Science and Engineering, the Guangdong Province Key Laboratory of Information Security Technology, and the Key Laboratory of Machine Intelligence and Advanced Computing, Ministry of Education, Sun Yat-sen University, Guangzhou 510006, China (e-mail: zhangx227@mail2.sysu.edu.cn; xiexiaoh6@mail.sysu.edu.cn; stsljh@mail.sysu.edu.cn; zhwshi@mail.sysu.edu.cn).}}
\maketitle

\begin{abstract}
We are concerned with retrieving a query person from multiple videos captured by a non-overlapping camera network. Existing methods often rely on purely visual matching or consider temporal constraints but ignore the spatial information of the camera network. To address this issue, we propose a pedestrian retrieval framework based on cross-camera trajectory generation, which integrates both temporal and spatial information. To obtain pedestrian trajectories, we propose a novel cross-camera spatio-temporal model that integrates pedestrians' walking habits and the path layout between cameras to form a joint probability distribution. Such a spatio-temporal model among a camera network can be specified using sparsely sampled pedestrian data. Based on the spatio-temporal model, cross-camera trajectories can be extracted by the conditional random field model and further optimized by restricted non-negative matrix factorization. Finally, a trajectory re-ranking technique is proposed to improve the pedestrian retrieval results. To verify the effectiveness of our method, we construct the first cross-camera pedestrian trajectory dataset, the Person Trajectory Dataset, in real surveillance scenarios. Extensive experiments verify the effectiveness and robustness of the proposed method.
    \end{abstract}


    \section{Introduction}

    The frame images, video tracklets, and cross-camera trajectories of a pedestrian form hierarchical clustering can be regarded as the low-order representation to the high-order representation of the pedestrian progressively. Compared to a single frame image, the clusters generally provide richer information. In other words, using person tracklets or person trajectories can enhance the person retrieval. However, how to effectively obtaining the person trajectory is still an open question. The existing pedestrian detection and intra-camera pedestrian tracking technologies can get the set of person tracklets under each camera well. In our method, we further infer the cross-camera trajectories of a pedestrian by combining the tracklet matching, conditional random field, and nonnegative matrix factorization technologies. After that, we develop a trajectory re-ranking technology to optimize the person retrieval results.
    Note that in our method the query may be a pedestrian tracklet or only a single pedestrian image, without the need of the query spatio-temporal information.

    A key to cross-camera pedestrian analysis is to model the empirical spatio-temporal relationship of the camera network. However, spatio-temporal modeling is challenging. Firstly, the pedestrian trajectories are often varying. For example, there may be multiple paths between a pair of cameras, and the walking time of pedestrians on the same way may also be different. Secondly, there is often a lack of data for establishing the spatio-temporal model. Currently, there are mainly three datasets can support the research of exploiting the spatio-temporal relationship between cameras for person retrieval, i.e., Market1501 \cite{Market1501}, DukeMTMC \cite{DukeMTMC}, and Campus4k \cite{xie2020progressive}. However, these datasets only contain the images or video frames with their shooting time but no additional information such as the distance between cameras. Based on these databases, some researchers have attempted to use temporal constraints to improve the accuracy of person association \cite{huang2016camera, lv2018unsupervised, wang2019spatial}. These methods mainly consider the distribution of time intervals when the same person appears in a pair of cameras. In practice, it is difficult to collect enough samples for the distribution statistics of the time interval between each camera pair. To overcome this shortcoming, we propose a new spatio-temporal model, which generates a joint probability distribution of spatio-temporal information by integrating pedestrians' walking habits and path distances between cameras. Here, walking habits are mainly related to walking speed and path selection propensity between cameras. The model exploits not only visual cues but also non-visual cues, including camera coordinates and path maps of regions of interest, which are readily available in real scenes. With the help of non-visual cues, the complete spatio-temporal relationship among the entire camera network can be modeled by sparse sampling of pedestrian visual data. Based on the spatiotemporal model, we develop a conditional random field (CRF) model to infer the cross-camera spatiotemporal adjacency graph. Finally, we propose to optimize pedestrian trajectories from spatiotemporal adjacency graphs via restricted nonnegative matrix factorization.

In order to verify the effectiveness of the proposed method, we collected a cross-camera pedestrian trajectory dataset in a real residential areas for experiments. Experiments show that the inference of pedestrian trajectories improves the accuracy of pedestrian retrieval. Meanwhile, experiments show that our method can also be combined with other Bayesian-based spatio-temporal methods to achieve better results.

      The contributions of this paper are as follows:

    \begin{itemize}
        \item A novel framework for pedestrian retrieval based on cross-camera trajectories is proposed. Pedestrian trajectories implying temporal and spatial information of the camera network can help improve the accuracy of pedestrian retrieval and further support other applications. In particular, in our framework, we do not need the spatio-temporal information of the query.
        \item A joint probability distribution based spatiotemporal model is designed to support trajectory generation in camera networks, which embeds pedestrian walking habits and path distances between cameras, and can be constructed by sparse sampling of pedestrian data.
        \item A spatiotemporal conditional random field method is developed to infer pedestrian trajectories. A non-negative matrix factorization is employed to optimize the inference results. And a trajectory re-ranking technique is proposed to improve pedestrian retrieval results.
        \item The first cross-camera person trajectory dataset from the real scene is collected. An evaluation method for person trajectory retrieval is also introduced. The dataset and code are released at https://github.com/zhangxin1995/PTD.
        
    \end{itemize}

    The remainder of this paper is organized as follows. The related works are reviewed in Section \ref{sec:related}. Our dataset and the evaluation method of trajectory generation are introduced in Section \ref{Dataset}. The proposed method is introduced in Section \ref{sec:method}. Experimental results are shown in Section \ref{sec:experiment}. Conclusion is in Section \ref{sec:Conclusion}.

    \section{Related work}\label{sec:related}
    This section reviews related works about spatio-temporal modeling, trajectory generation, and re-ranking, which are all critical parts of our proposed framework.

    \subsection{Spatio-Temporal Model for Pedestrian Matching}
    To improve the accuracy of pedestrian matching, some researchers have proposed combining spatio-temporal information with visual features. For this purpose, how to build the spatial-temporal model for a camera network is essential. In 2013, Xu et al.\cite{6589140} proposed a graph model that captures the relationships among objects for target retrieval in the distributed camera network, which does not identify global trajectories. In 2018, Lv et al.\cite{lv2018unsupervised} proposed using spatio-temporal constraints in their unsupervised person Re-ID method. They make a strong assumption that a gallery person always appears in the time window $(t-\bigtriangleup t,t + \bigtriangleup  t)$ when given a query person image at the time $t$. In 2020, Xie et al.\cite{xie2020progressive} proposed an association method to integrate both spatio-temporal probability and visual probability into a joint probability through an unsupervised way, which is to optimize person retrieval results in the prediction phase. In 2016, Huang et al.\cite{huang2016camera} proposed using Weibull distribution to model the distribution of cameras pair and to use it in the sorting stage after searching. In 2019, our group \cite{wang2019spatial} proposed a fast Histogram-Parzen (HP) method to fit the spatio-temporal relationship between cameras. Above mentioned methods all focus on the relationship between two cameras while ignoring the overall relationship among a camera network. Therefore, these methods require sufficient data between each camera pair to establish a complete spatio-temporal relationship, which is difficult to meet in practice. In the field of vehicle re-identification, Liu et al.\cite{8036238} propose a method that incorporates spatial-temporal and appearance information, where inversely proportional functions model the temporal and spatial relationships separately.
In this paper, we model the spatio-temporal relationship of the camera network as a whole, which embeds the path distances between cameras. Based on this model, we can complete the construction of the spatio-temporal relationship between all camera pairs with a sparse sampling of pedestrian spatio-temporal trajectory. In addition, the existing person retrieval methods using spatio-temporal information are mainly based on the Bayesian strategy, which requires knowing the spatio-temporal information of the query in advance, which limits the use scenario of this kind of method. The clustering and scattering mechanism proposed in this paper can optimize the retrieval when the query spatio-temporal information is unknown, and is applicable to a wider range of scenarios.

    \subsection{Pedestrian Trajectory Generation}
   
    Some existing video analysis tasks have considered the generation of cross-camera pedestrian trajectories. For example, Multi-Target Multi-Camera Tracking(MTMCT) aims to determine the positions of every person at all times from video streams taken by multiple cameras \cite{LAAM}. According to the different application scenarios, it can be divided into centralized and distributed tracking approaches. The centralized approach is wildly used in overlapping camera view tracking systems. It aims to reduce the effect of occlusions and noisy observations in a crowded environment or in the monitoring of small areas \cite{liem2014joint,nie2014single}. The distributed approach is generally used in applications which are dealing with non-overlapping camera views that are designed to monitor large areas most of the time \cite{iguernaissi2019people}. In this kind of MTMCT method, they usually assume that the camera's field of view is often non-overlapping and there is no explicit geometric relationship between cameras. Javed et al. \cite{javed2008modeling} proposed a method that calculates the similarity of the appearance and that of spatio-temporal data between every two cameras to judge who this person is.
    The difference between the above work and ours is that MTMCT focused on the association, but ours focused on both association and retrieval collaboration. Furthermore, existing MTMCT studies have limitations, such as small intervals between cameras, short duration in a video, and fewer tracking targets. Our work exceeds these limits.

    \subsection{Re-ranking for Person Retrieval}
    Re-ranking means using high confidence samples to reorder the initial search results, which has been studied in the generic instance retrieval  \cite{zhong2017re,chum2007total,jegou2007a,qin2011hello,qin2011hello,shen2012object}. Some researchers have also designed re-ranking methods for person Re-ID task \cite{zheng2015query,chen2016deep,garcia2015person,garcia2017discriminant,li2012common}. Li et al. \cite{li2012common} proposed a re-ranking method by analyzing the relative information and direct information of near neighbors of each pair of images. Garcia et al. \cite{garcia2015person} proposed an unsupervised re-ranking method by jointly considering the content and context information in the ranking list. Leng et al. \cite{2015Person} proposed a bidirectional ranking method to modify the initial ranking list based on both content and context similarity. Zhong et al. \cite{zhong2017re} proposed a k-reciprocal encoding method to re-ranking the Re-ID results. These re-ranking methods only use the visual information of persons while ignoring the temporal or spatial information. In 2019, our group \cite{wang2019spatial} proposed a spatio-temporal two-stream re-ranking method to reorder the retrieval result list. However, this method requires the query's spatio-temporal information to be known, limiting its application scenarios. In this paper, we develop a novel trajectory re-ranking technology to optimize person retrieval. The proposed trajectory re-ranking method doesn't need the spatio-temporal information of the query.

    \begin{figure}[t]
        \centering
        \includegraphics[width=0.4\textwidth]{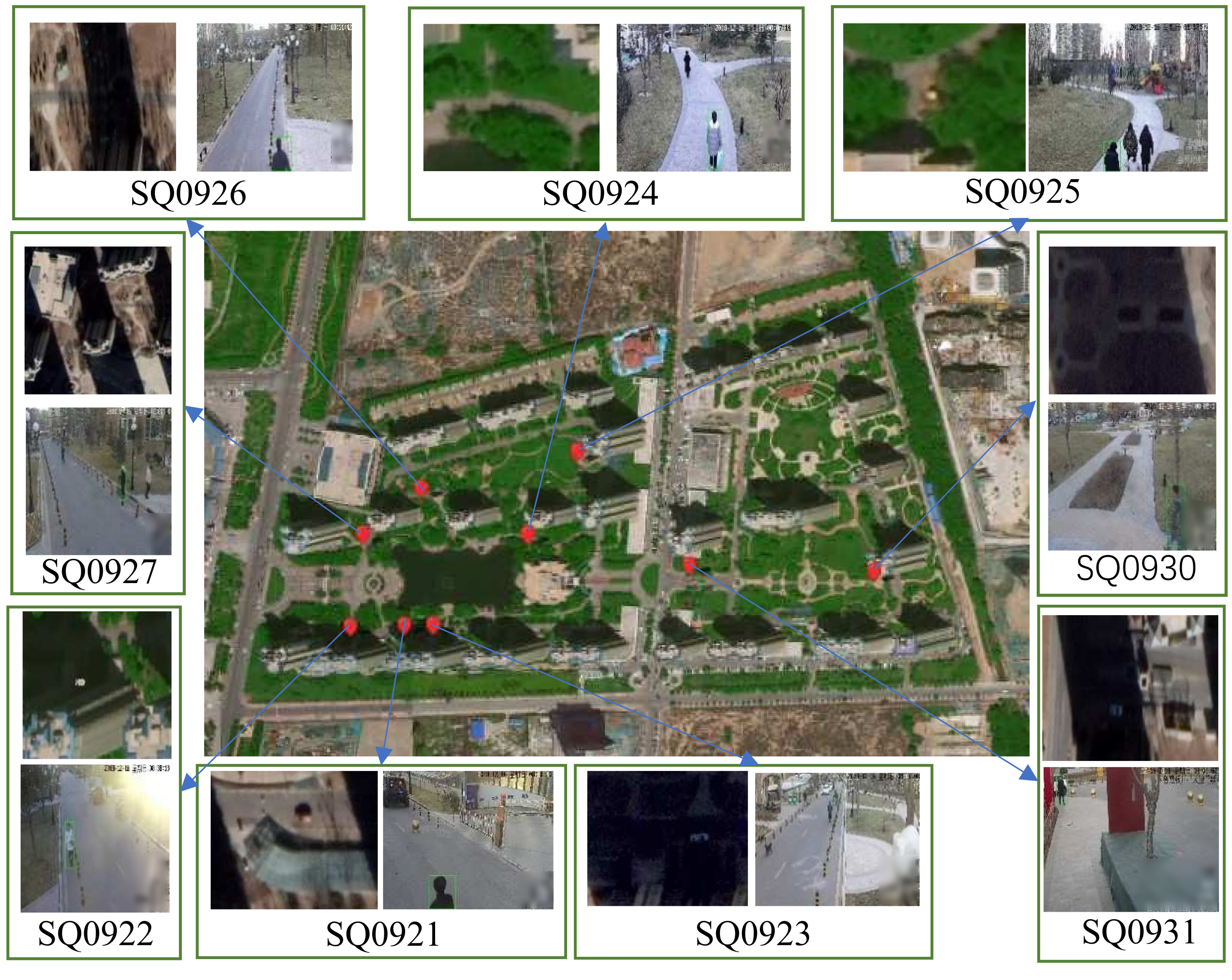}
        \caption{The spatial distribution of the cameras in the Person Trajectory Dataset. For each camera, the satellite's enlarged image and the camera view of the corresponding cameras are displayed. The following numbers, such as SQ0921, indicate the camera index.}
        \label{fig:SDCPTD}
    \end{figure}
    
    \section{Person Trajectory Dataset}\label{Dataset}

    Thanks to many public Re-ID datasets \cite{Market1501,DukeMTMC,xie2020progressive,zheng2016mars}, great progress has been made in person retrieval. However, existing Re-id datasets do not provide labels of the cross-camera trajectory of persons. This brings a great challenge to the research of spatio-temporal relationship. To break this dilemma, we collect a dataset called Person Trajectory Dataset (PTD), containing complete image sequences from non-overlapping outdoor cameras in a residential area, labeled full person trajectories.

    \begin{figure*}[t]
        \centering
        \subfloat[]{\includegraphics[width=0.33\textwidth]{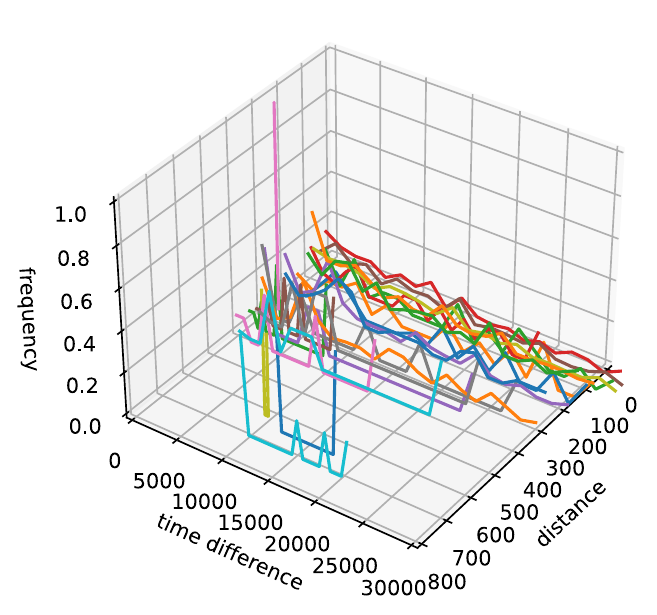}\label{fig3a}}
        \subfloat[]{\includegraphics[width=0.33\textwidth]{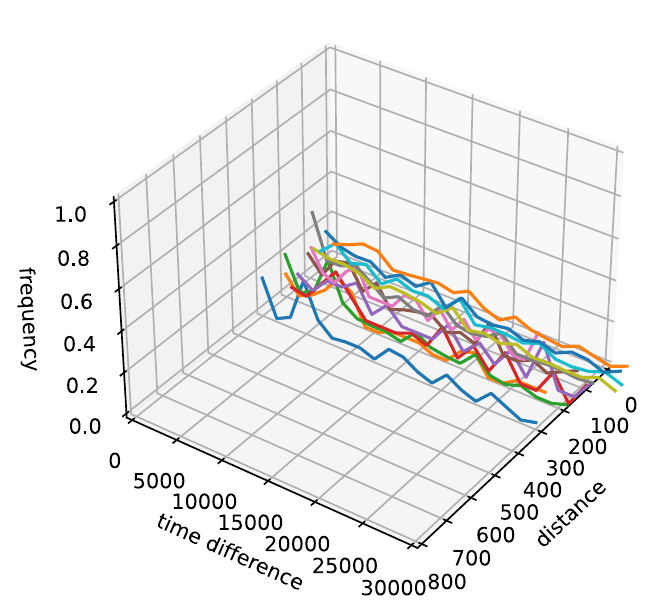}\label{fig3b}}
        \subfloat[]{\includegraphics[width=0.33\textwidth]{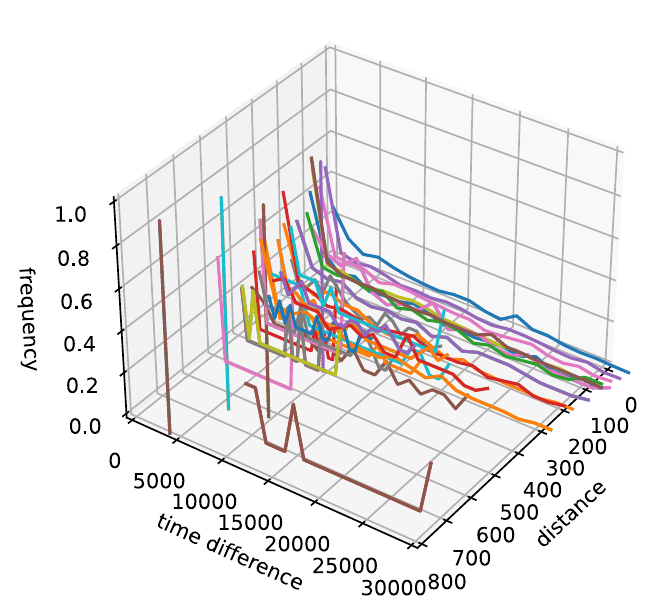}\label{fig3c}}
       \caption{Spatio-temporal distribution under different camera pairs in Spatio-Temporal Dataset. (a) Spatio-temporal distribution of positive samples in the training set. (b)The spatio-temporal distribution of positive samples selected for training. (c)Spatio-temporal distribution of positive samples in the testing set. Curves with different colors represent different camera pairs, and the Z-axis represents the frequency of occurrence under the corresponding distance and time difference.}
       \label{dist}
    \end{figure*}
    \begin{figure}[t]
        \includegraphics[width=0.5\textwidth]{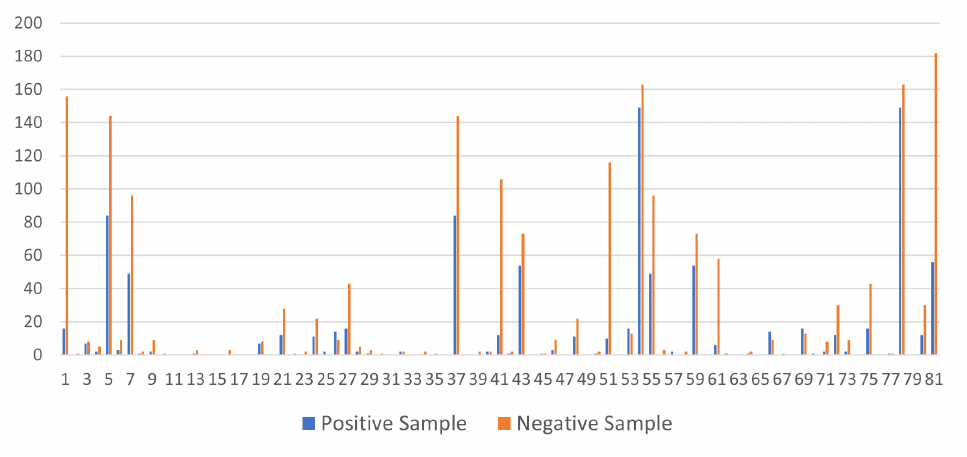}
        \caption{ The ``positive" and ``negative" samples in the training set of PTD. The horizontal axis is the index of 81 pairings between 9 cameras. The vertical axis is the number of positive and negative samples in each pair of cameras.}
        \label{Fig:PN}
    \end{figure}

    This PTD is collected from a camera network of nine cameras from 8 a.m. to 8 p.m. The spatial distribution of the cameras for this dataset is shown in Fig. \ref{fig:SDCPTD}. The person image sequence acquired is preliminarily labeled by the FPN target detection method \cite{8099589} and DeepSort tracking method \cite{Bewley2016_sort}, then the annotation results are corrected manually. The images of any person that only appears under one camera are also eliminated to ensure the persons in PTD have occurred under at least two different cameras.

    The dataset can be divided into two parts: visual data and spatio-temporal data. The visual data includes 17,996 images of 662 individuals, of which 5,003 images of 197 individuals are used as the testing set, and the remaining 12,993 images of 465 people are used as the training set.

    Spatio-temporal data also consists of a training set and testing set, which are derived from PTD's training set and testing set, respectively. Fig. \ref{Fig:PN} shows the amount of data between each camera pair in spatio-temporal data. From Fig. \ref{Fig:PN}, we can see that it is tough to establish the spatio-temporal relationship model of each camera pair when there is limited data. First, there is an imbalance between different camera pairs. For example, there are 228 samples in camera pair No. 3, while there are only 12 samples in camera pair No. 6. Second, the number of positive and negative samples is also unbalanced across the camera pair. In spatio-temporal data, a positive example refers to the time difference in a specific person's trajectory under each camera. A negative example refers to the time difference between the different trajectories of the same person and other people under all cameras. The sample imbalance between camera pairs poses a great challenge to the spatio-temporal modeling between cameras. Because of the imbalance in spatio-temporal data, only ten camera pairs are used in training. Fig. \ref{dist} shows the temporal distribution under different camera pairs in Spatio-Temporal Dataset. In Fig. \ref{dist}, the distance represents the shortest path length between the corresponding camera pairs, and the temporal difference means the absolute value of the time past between a person's appearances in their corresponding camera pair. From Fig. \ref{fig3a} and Fig. \ref{fig3c}, we can see that the spatio-temporal distribution between the training set and the test set is significantly different. Fig. \ref{fig3b} shows the spatio-temporal distribution of camera pairs selected for training from the training set. Compared with Fig. \ref{fig3a}, Fig. \ref{fig3b} removes some camera pairs with very little data. The figure shows that the data distribution we used for training is very sparse, so it is challenging to mine the spatio-temporal information.

    We want to compare the PTD with the existing MTMCT datasets. There are mainly three public MTMCT benchmark datasets, PET09S2L1 \cite{ferryman2010pets2009}, CAMPUS \cite{xu2016multi}, and EPFL \cite{4359319}. In PETS, seven cameras are used to film about ten targets entering and walking through a footpath. In CAMPUS, they collected about 15 persons in the four video subsets. The amounts of persons in these MTMCT datasets are too small to be used to evaluate person retrieval tasks. Compared with these MTMCT datasets, the PTD data comes from the surveillance camera of the real scene, with more extensive coverage and more pedestrians recorded.

    \section{Proposed Method}
    \label{sec:method}
    
    \subsection{Overview}
    \begin{figure*}[t!]
        \includegraphics[width=\textwidth]{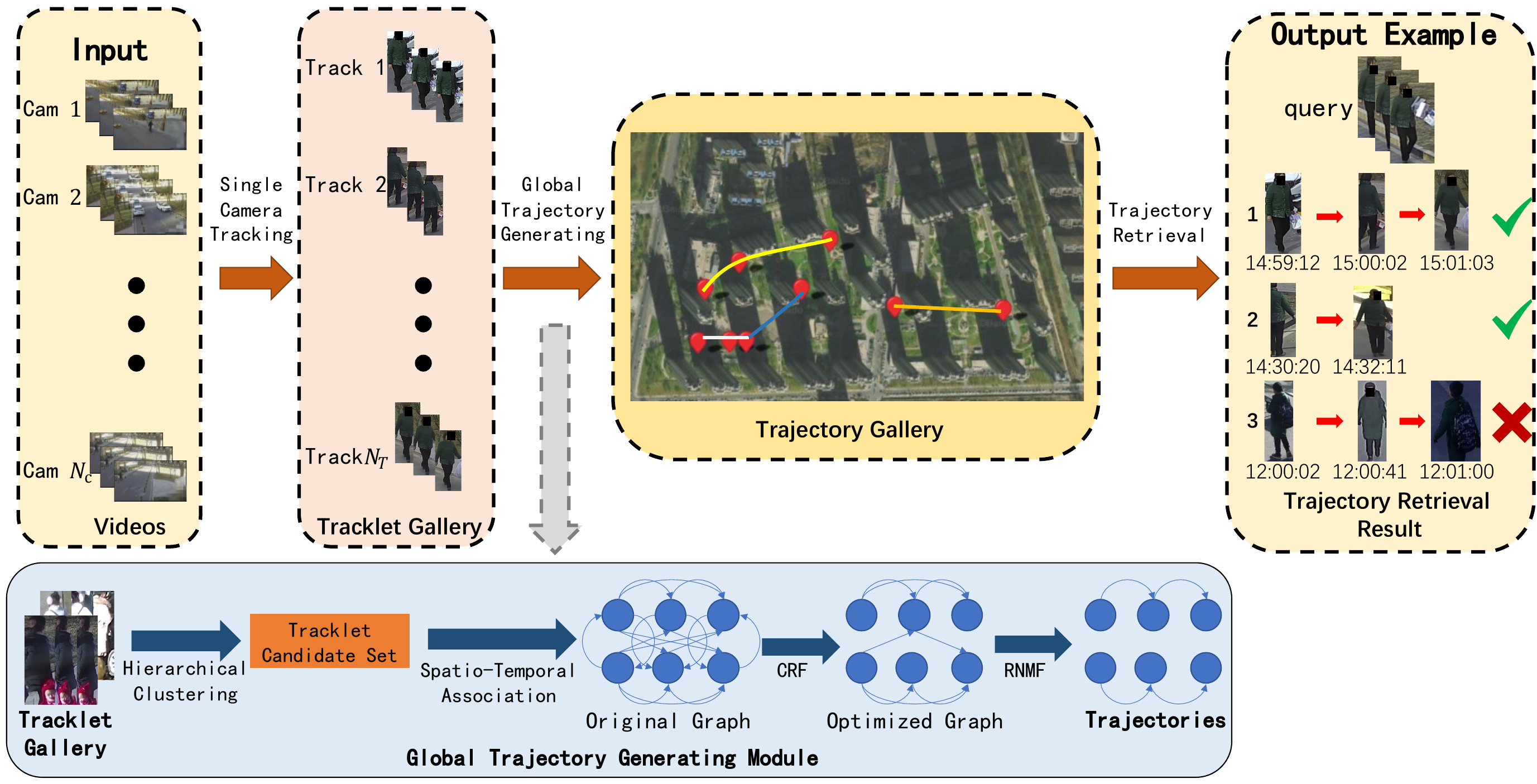}
        \caption{The overview of the person retrieval task and the framework of our method for trajectory generation. A green tick indicates that the trajectory identity is the same as the query, and a red cross indicates that the trajectory identity differs from the query. In output example, each photo represents represents the first image of a tracklet. }
        \label{fig:framework}
    \end{figure*}
    \begin{table}[t]
        \caption{GLOBAL VARIABLE DEFINITION}
        \begin{tabularx}{0.5\textwidth}{p{0.03\textwidth} X}
            \toprule
            \underline{\textbf{\emph{Variables}}} \\
            $\alpha_k$ &  Index of the $k$-th camera\\
            $N_{\alpha_k}$ & Number of tracklets under the camera $\alpha_k$\\
            $N_c$ & Number of cameras \\
            $T$ & Set of all tracklets in gallery\\
            $N_T$ & Number of tracklets in $T$\\
            $T^{\alpha_k}$ & Set of tracklets under the camera $\alpha_k$ \\
            $c_i$ & Index of the camera to which the $i$-th tracklet in $T$ belongs\\
            $t_i$ & Timestamp of the $i$-th tracklet in $T$ appearing under the camera $c_i$\\
            $d_{ij}$ & Shortest distance between camera $c_i$ and $c_j$ \\
            $t_{ij}$ & Time difference between $i$-th tracklet and $j$-th tracklet in $T$ \\
            $o_i^{\alpha_k}$ & $i$-th tracklet under the camera $\alpha_k$\\
            $N_{o_i^{\alpha_k}}$ & Number of images of tracklet $o_i^{\alpha_k}$\\
            $I_{c_i,t_i}$ & One image of $i$-th tracklet in $T$ shot by the camera $c_i$ at time $t_i$\\
            $I^{\alpha_k}_{ij}$ & $j$-th image of the $i$-th tracklet under the camera $\alpha_k$\\
            $f_i^{\alpha_k}$ & Feature vector of the $i$-th tracklet under the camera $\alpha_k$\\
            $w_m$ & $m$-th class in the clustering result\\
            $\hat{w}$ & Set of clustering results consisting of all classes\\
            $f_{query}$ & Feature vector of input query \\
            $S$ & Generated trajectory gallery \\
            $N_S$ &Number of trajectories in S \\
            $S_i$ & $i$-th trajectory in $S$ \\
            $N_{S_i}$ & Number of tracklets in $S_i$ \\
            $f_{S_i}$ & Feature vector of $S_i$\\
            $z_i$ & Given a query, $z_i$ represents the index in $S$ of the $i$-th trajectory retrieval result. \\
            $d_{S_{z_i}}$ & Cosine distance between the query and $S_{z_i}$.\\
            $sim^*_{S_{z_i}}$ & Similarity between the query and $S_{z_i}$ after query spatio-temporal re-ranking.\\
            $n_{o^{\alpha_k}_{i}}$ & Number of occurrences of $o^{\alpha_k}_i$ in $S$ \\
            $\eta_{S_{z_i}}$ & Spatio-temporal probability between the query and the trajectory $S_{z_i}$\\
            \bottomrule
        \end{tabularx}
        \label{SDT}
    \end{table}

    In this paper, we mainly solve three sub-problems. The first is building the spatio-temporal model under a camera network. The second one is how to use the specific spatio-temporal model to extract the possible trajectories of different persons. The third one is how to retrieve the trajectories according to the query. All global variables we will use are defined in Table \ref{SDT}, and other local variables are explained further later in the paper. The spatio-temporal relationship between cameras can be defined as a function $\psi(I_{c_i,t_i}, I_{c_j,t_j})$, where $\psi$ is a function that outputs the probability of $I_{c_i,t_i}$ and $I_{c_j,t_j}$ belonging to the same trajectory. We need to specify the spatio-temporal model $\psi$ by training samples between different cameras and predict the spatio-temporal relationship of testing samples. For the trajectory generation model, the input includes the query image and a gallery, and the output is a trajectory set. Each trajectory should satisfy two constraints: similarity constraint and spatio-temporal constraint. Similarity constraint means that the query and the images of the generated trajectory should be similar. The spatio-temporal constraint means that the generated trajectory should satisfy the spatio-temporal relationship between cameras. The goal of the person retrieval with trajectory generation is to retrieve all possible trajectories of the query person. The overview of the person retrieval and the framework of our method are shown in Fig. \ref{fig:framework}. The person retrieval task consists of three steps. The first step is data preprocessing, which uses the intra-camera tracking algorithm to extract the pedestrian tracklets in the video to form a tracklet gallery. Next, we use the global trajectory generation algorithm to transform the intra-camera tracklets into cross-camera trajectories. The third step is to give one or more images of a target person, and return the trajectories of the person with the same identity as the target person in the global trajectory gallery. Our work mainly focuses on the cross-camera trajectory generation stage. In short, our method mainly includes the following steps:
    
    \begin{enumerate}
        \item  Constructing a cross-camera spatio-temporal model.
        \item  Obtaining the visually similar candidate tracklets from the gallery set according to pure visual cues.
        \item  Using the spatio-temporal model to obtain the spatio-temporal probability between different tracklets in the candidate set.
        \item Using the conditional random field method to obtain a spatio-temporal consistent graph model.
        \item Using non-negative matrix factorization to infer the trajectories in the optimized graph model.
        \item Using trajectory re-ranking technology to enhance pedestrian retrieval.
    \end{enumerate}

    \subsection{Spatio-Temporal Modeling}
    \label{section:spatio-temporal}
    We propose a novel spatio-temporal model based on the path distances between camera pairs. We assume that the camera network in an area satisfies the spatio-temporal relation $\psi(d_{ij},t_{ij})$. The $\psi(d_{ij},t_{ij})$ is a two-dimensional probability distribution function. When the path distance between two cameras and the corresponding time length are input into the model, it returns a probability to indicate the likelihood that the person is moving continuously from one camera to another. Because we often do not have enough data for all camera pairs to compute the spatio-temporal probability distribution, we choose to establish a joint probability model of the camera network. Firstly, we train the spatio-temporal model using the camera pairs with enough data and then get the spatio-temporal probability of all camera pairs by interpolation. We use Multi-Layer Perceptron (MLP) to fit the spatio-temporal probability model $\psi$. MLP is smooth and continuous, which satisfies the assumption that the spatio-temporal model is a two-dimensional continuous probability distribution. Our model consists of three network layers: input, hidden, and output. The input layer inputs the path distance $d_{ij}$ and time length $t_{ij}$. The output layer returns whether the input value meets the time constraint. Assuming that the probability of the final output is $\hat{y_i}$ and the label is $y_i$, we train the model with the cross entropy loss function, which is expressed by the mathematical formula as follows:
    \begin{equation}
        Loss=\sum_{i=1}^n y_ilog(\hat{y_i})
    \end{equation}
    where $n$ is number of mini-batch.

\begin{figure*}[htp]
    \centering
        \subfloat[]{\includegraphics[width=0.3\textwidth]{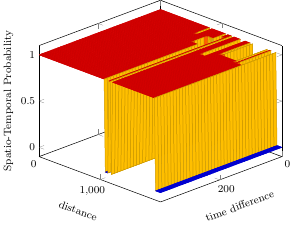}\label{fig:decisiontree}}
        \subfloat[]{\includegraphics[width=0.3\textwidth]{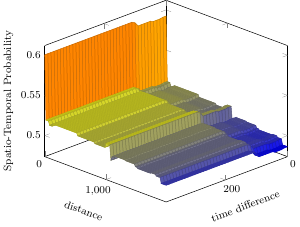}\label{fig:adaboost}}
        \subfloat[]{\includegraphics[width=0.3\textwidth]{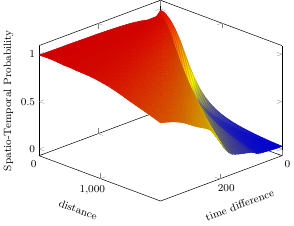}\label{fig:MLP}}
        \caption{Visualization of spatio-temporal fitting by different models. (a) DecisionTree. (b) Adaboost. (c) MLP. Their AUC scores are 0.934, 0.986, and 0.987, respectively, on the test set of PTD.}
        \label{fig:Visualization}
    \end{figure*}

    Fig. \ref{fig:Visualization} illustrates the spatio-temporal fitting on PTD using different models, including DecisionTree,  Adaboost, and MLP. As shown, the probability distribution obtained by MLP is more smooth, which is more in line with the actual situation of pedestrians walking. In the experiment section, we will verify that MLP indeed works better for the final pedestrian retrieval.

    \subsection{Extraction of Candidate Tracklet Set}
    We use visual similarity to obtain the candidate tracklet set. Hierarchical clustering(HC) is a common clustering method that doesn't need to specify the number of cluster categories. We define $T$ as the set of all tracklets in the gallery. And we represent a set of tracklets under one camera by $T^{\alpha_k}=\{o^{\alpha_k}_1,o^{\alpha_k}_2,...,o^{\alpha_k}_{N_{\alpha_k}}\}$ where $\alpha_k$ is the index of the $k$-th camera, $o^{\alpha_k}_i$ is the $i$-th tracklet under the camera $\alpha_k$ and $N_{\alpha_k}$ is the number of tracklets under the camera $\alpha_k$. We use a feature extraction model, such as Resnet50 \cite{7780459}, to extract features for each tracklet in the gallery, which can be expressed as follows:
    \begin{equation}
       f^{\alpha_k}_i=\Omega(o^{\alpha_k}_i)=\frac{\sum_{j=1}^{N_{o_i^{\alpha_k}}}F(I^{\alpha_k}_{ij})}{N_{o_i^{\alpha_k}}}
    \end{equation}
where $\Omega$ is the average function, $F$ is the image feature extract function and $I^{\alpha_k}_{ij}$ is the $j$-th image of the $i$-th tracklet under the camera $\alpha_k$. By the feature extraction model, we can get the feature set of the tracklet gallery. For single camera $\alpha_k$, the features of $i$-th tracklet are extracted and averaged to get the corresponding feature vector $f_{i}^{\alpha_k}$. Then, the hierarchical clustering algorithm is carried on the set $\omega=\{f^{\alpha_1}_{1},f^{\alpha_1}_{2},...,f^{\alpha_{N_c}}_{N_{\alpha_{N_c}}}\}$ to get the clustering result $\hat{\omega}=\{\omega_{1},\omega_{2},\cdots,\omega_{m}\}$, where $\omega_{m}=\{f^{\alpha_k}_{i} \lvert o^{\alpha_k}_{i} \in class\,m\}$ represents the feature set of the $m$-th class in the clustering result.

    \subsection{Cross-Camera Trajectory Generation}
        In this section, we introduce our trajectory generation module, which can be divided into two stages, adjacency graph generation and trajectory optimization.

    \subsubsection{Adjacency Graph Generation}
    At this stage, we aim to get the spatio-temporal probability between the members in the candidate set. We take the candidate set generated by HC as the input. First, we define that each element $f^{\alpha_k}_i$ in the candidate set $\omega_m$ is a vertex in the graph and all the vertices constitute the vertex set $V=\{v_1,v_2,...,v_{|w_m|}\}$. We define the index of vertex $v_p$ in $T$ as $\beta_p$ and define the $e_{pq}$ as the edge between $v_p$ and $v_q$. The direction of $e_{pq}$ is determined by the time of $v_p$ and $v_q$. When the time $t_p$ is before $t_q$, the direction is from $v_p$ to $v_q$, otherwise from $v_q$ to $v_p$. All edges $e_{pq}$ form the set $E$. Then we define a probability function $Q(e_{pq})$ whose inputs are edges $e_{pq}$ in $E$. The $Q$ function can be expressed by the following formula:
        \begin{equation}
                  Q(e_{pq})=\psi(d_{\beta_{p}\beta_{q}}, t_{\beta_{p}\beta_{q}})
        \end{equation}
        where $\psi$ is the spatio-temporal model introduced in Section \ref{section:spatio-temporal}. After that, then we apply the conditional random field to obtain a spatial-temporal consistent trajectory model. The spatial-temporal association graph obtained from the previous step is defined as the graph $G=\{V,E\}$. The graph obtained after conditional random field processing is represented by $\hat{G}=\{V,\hat{E}\}$. And the total probability model of the spatio-temporal consistency graph of the trajectory is defined as $\Pi(V,E)$. Therefore, the method of obtaining spatio-temporal consistency graph can be expressed by the following formula:
        \begin{equation}
        \label{ai}
                    \hat{E}=\mathop{\arg\max}\limits_{E}\Pi(V,E)
        \end{equation}
        The total probability $\Pi(V,E)$ describes the probability that the constructed graph satisfies spatial-temporal consistency under the condition that the edges are $E$ and the nodes are $V$. When the total probability $\Pi(V,E)$ obtains the maximum value, the corresponding $E$ is the solution $\hat{E}$. Because the total probability formula is difficult to calculate, we solve the formula (\ref{ai}) by approximate inference. We assume that the probability of each pair of nodes satisfying spatial-temporal constraints is independent of each other, then the total probability formula can be expressed as:
        \begin{equation}
        \label{ve}
            \Pi(V,E)\approx\prod{\Lambda(e_{pq})}
        \end{equation}
where $\Lambda(e_{pq})$ represents the probability of whether the edge satisfies spatial-temporal consistency. We use the mean field method to solve the $\Lambda(e_{pq})$. By using the mean field method, the $\Lambda(e_{pq})$ can be calculated iteratively by the following formula:
        \begin{equation}
        \label{energy}
            \Lambda^{P+1}(e_{pq})=\frac{e^{\iota[\rho_1 (\Lambda^{P}(e_{pq})-u_1)+\rho_2(\Upsilon(v_p,v_q)-u_2)]}}{Z}
        \end{equation}
where $\rho_1, \rho_2, u_1, u_2$ are hyper-parameters, $P$ represents the number of current iteration. For the first iteration, we initialize $\Lambda^{0}(e_{pq})=Q(e_{pq})$. $Z$ is the normalization term, and the following formula can calculate it:
        \begin{equation}
            Z=e^{\iota[\rho_1 (1-u_1)+\rho_2(1-u_2)]}
        \end{equation}
        $\Upsilon(v_p,v_q)$ is the spatial-temporal consistent probability of vertex $v_p$ and vertex $v_q$, it can be calculted by following formula:
        \begin{equation}
            \Upsilon(v_p,v_q)=\frac{\sum_{s}{\Lambda^P(e_{ps})\Lambda^P(e_{qs})}}{\sqrt{\sum_{s}{\Lambda^P(e_{ps})^2}}\sqrt{\sum_{s}{\Lambda^P(e_{qs})^2}}}
            \label{f:upsilon}
        \end{equation}
        Then we reset each $\Upsilon^{P+1}(e_{pp})$ to 1, because the spatio-temporal consistency between themself is 1. After $P^*
        $ iterations, we can get the $\Lambda^{P^*}(E)$. $\Lambda^{P^*}(E)$ is the set of all $\Lambda^{P^*}(e_{pq})$. Finally, we set a threshold $\kappa$ and remove all the edges less than $\kappa$ in $\Lambda^{P^*}(E)$, and the remaining edges are set $\hat{E}$.

    \subsubsection{Trajectory optimization}
     We model the problem of extracting pedestrian trajectory from graph $\hat{G}(V,\hat{E})$ as the following mathematical problem: Given graph $\hat{G}(V,\hat{E})$, we need to find one or more subsets of it, which satisfy the following rules:
     
    \begin{enumerate}[i)]
        \item Each subset represents a cross-camera trajectory. The nodes in the trajectory are sorted according to time and the time length that the nodes should meet a certain spatio-temporal relationship.
        \item Each subset includes at least one node.
        \item The union of all subsets is equal to $V$.
        \item Each node from $V$ appears in only one subset.
    \end{enumerate}

     We choose restricted non-negative matrix factorization(RNMF)\cite{9042858} to get the final pedestrian trajectory. Using RNMF has two advantages in this task. One is that it does not need to know the number of clusters in advance. The other is that its performance is the best compared with other algorithms. First, we represent graph $\hat{G}(V,\hat{E})$ as adjacency matrix $W$, its size is $ N_V \times N_V $. $N_V$ is the number of nodes in set $V$, and the element $W_{ij}$ in $W$ is weight of $\hat{e}_{ij}$ from $\hat{E}$. Given the adjacency matrix $W$, we need to find an assignment matrix $A^*$ to satisfy the following constraints:
    \begin{equation}
    \label{fsa}
    \begin{aligned}
        A^*&=\mathop{\arg\min}\limits_{A}||W-AA^T||\\
        s.t. A &\in \{0,1\}^{N\times K} \\
        A1_k&=1_N
    \end{aligned}
    \end{equation}
where $A$ is the assignment matrix, a binary matrix whose size is $N_V \times K$. $K$ is the number of possible trajectories in $W$. To obtain the value of $K$, we calculate the eigenvalues of S and define the absolute values of all eigenvalues as $\varsigma_1,\varsigma_2,...,\varsigma_n$ . We define a threshold $\varpi$ and get the value of K through the following formula:
    \begin{equation}
        K=\sum_i{\mathbb{I}(\varsigma_i>\varpi)}
    \end{equation}
where $\mathbb{I}(\varsigma_i>\varpi)$ is a function that when $\varsigma_i>\varpi$ is true, it is 1, otherwise it is 0.
    The $A$ indicates which trajectories each node should be assigned to. The $1_N$ and $1_K$ are defined as the column vector of $N\times 1$ and $K\times 1$ with value of 1. Under the condition of binary, equation (\ref{fsa}) is difficult to solve, so we convert equation equation (\ref{fsa}) into the following formula to solve it:
    \begin{equation}
    \begin{aligned}
        &A^*=\mathop{\arg\min}\limits_{A'\geq 0}||S-A'A'^T||+\tau||A1_k-1_N||
    \end{aligned}
    \end{equation}
where $A'$ is the real version of $A$, $\tau$ is the hyper-parameter, and $||A1_k-1_N||$ is the regularization term. We solve this equation iteratively according to \cite{9042858}:
    \begin{equation}
        A'^*=A'\odot \sqrt{(SA'+2\tau1_K1_N^T)\oslash (4A'A'^TA'+2\tau1_K1_K^T)}
     \end{equation}
where $\oslash$ represents the division of matrix elements and $\odot$ represents the multiplication of matrix elements. After obtaining the solution $A'^*$, we traverse all rows of $A'^*$ and select the column of the maximum value of each row as its corresponding trajectory.

 \subsection{Analysis of Conditional Random Field}

\begin{figure}
        \includegraphics[]{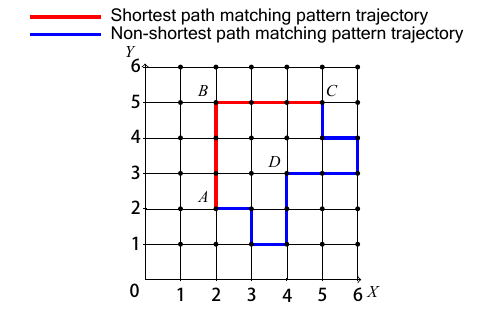}
        \caption{The diagram of two spatio-temporal matching pattern trajectories. Each dot represents a camera. The line segment between dots is the shortest path between two cameras. The line segment marked with red or blue represents a tracklet.}
        \label{f:example_trajectory_pattern}
    \end{figure}

The main reason for using conditional random field (CRF) modeling in cross-camera trajectory generation is that CRF can be compatible with two different spatio-temporal matching patterns. There are two spatio-temporal matching patterns for cross-camera trajectory as shown in Fig. \ref{f:example_trajectory_pattern}: One is the shortest path matching pattern. In this pattern, the shortest path of any two tracklets in a trajectory is always a sub-path of this trajectory. For example, in Fig. \ref{f:example_trajectory_pattern}, the red trajectory $AB$ is a sub-trajectory of the red trajectory $AC$, and the red $AB$ is the shortest path from point $A$ to point $B$. The spatio-temporal model built with the shortest path will make the spatio-temporal probability of any two tracklets in this trajectory as high as possible. The other is the non-shortest path matching pattern, in which the shortest path of any two tracklets in a trajectory is no longer always a sub-path of this trajectory. In Fig \ref{f:example_trajectory_pattern}, the blue trajectory $AD$ is the sub-trajectory of the blue trajectory $AC$, but the blue $AD$ isn't the sub-trajectory of the shortest path from $A$ to $C$. The spatio-temporal probability between adjacent tracklets in the trajectory is high, but the spatio-temporal probability between non-adjacent tracklets is low. The two spatio-temporal patterns pose a great challenge to getting the correct trajectory set. 

The CRF method is suitable for solving this problem because the CRF can transform the spatio-temporal matching pattern of some non-shortest paths into the shortest path matching pattern. Next, we theoretically prove that the conditional random field will simultaneously strengthen the spatio-temporal consistency of the two patterns' trajectories.

Given the initial spatio-temporal association graph $G(V,E)$, we use $e_{pq}$ to represent the edge between vertex $v_p$ and $v_q$, and their spatio-temporal consistency can be expressed by $Q(e_{pq})$. We initialize $\Lambda^{0}(e_{pq})=Q(e_{pq})$ and define the spatio-temporal probability gain function as follows:
\begin{equation}
    \label{f:gain}
    D(e_{pq},P)=\Lambda^{P+1}(e_{pq})-\Lambda^{P}(e_{pq})
\end{equation}
Therefore in the optimization process of the CRF, the increase of the spatio-temporal consistency of vertex $v_p$ and vertex $v_q$ is equivalent to inequality $D(e_{pq},P)>0$. Considering the first iteration, after the mathematical derivation in the appendix, $D(e_{pq},0)>0$ can be converted into the following inequality:
\begin{equation}
    \label{f:ieq3}
    \Gamma Q(e_{pq})-ln(Q(e_{pq}))>ln(Z)+\iota(\rho_1 u_1+\rho_2 u_2)-\pi
\end{equation}
where $\Gamma$ and $\pi$ are expressed by the following formulas:
\begin{equation}
    \label{f:approx1}
    \begin{aligned}
        \Gamma=\iota(\rho_1\!+\!\frac{\rho_2}{\sqrt{\sum_{s\neq p}{Q(e_{ps})^2}}\sqrt{\sum_{s}{Q(e_{qs})^2}}})
    \end{aligned}
\end{equation}
\begin{equation}
    \label{f:approx2}
    \begin{aligned}
    \pi=\iota\rho_2\frac{\sum_{s\neq p}{Q(e_{ps})Q(e_{qs})}}{{\sqrt{\sum_{s\neq j}{Q(e_{ps})^2}}\sqrt{\sum_{s}{Q(e_{qs})^2}}}}
    \end{aligned}
\end{equation}
For the shortest path pattern trajectory, $Q(e_{pq})$ is generally large, and it has better spatio-temporal consistency. Therefore, $\Gamma Q(e_{pq})-ln(Q(e_{pq}))$ and $\pi$ are both large, so the inequality $D(e_{pq},0)>0$ holds. In particular, even if some part of $Q (e_{pq})$ is not large enough because the spatio-temporal consistency of such trajectories is usually high, the CRF method can also strengthen spatio-temporal probability of such trajectories. For the trajectory of the non-shortest path pattern, the $Q(e_{pq})$ of the adjacent tracklet is high, but the $Q(e_{pq})$ of the non-adjacent tracklet is low. Therefore $\pi$ is also small and cause $ln(Z)+\iota(\rho_1 u_1+\rho_2 u_2)-\pi$ to rise. When strengthening such trajectory, CRF requires higher $Q(e_{pq})$ for adjacent tracklets than for the shortest path type trajectories, which can filter out more incorrect associations.

\begin{figure}[t!]
        \centering
        \includegraphics[width=0.4\textwidth]{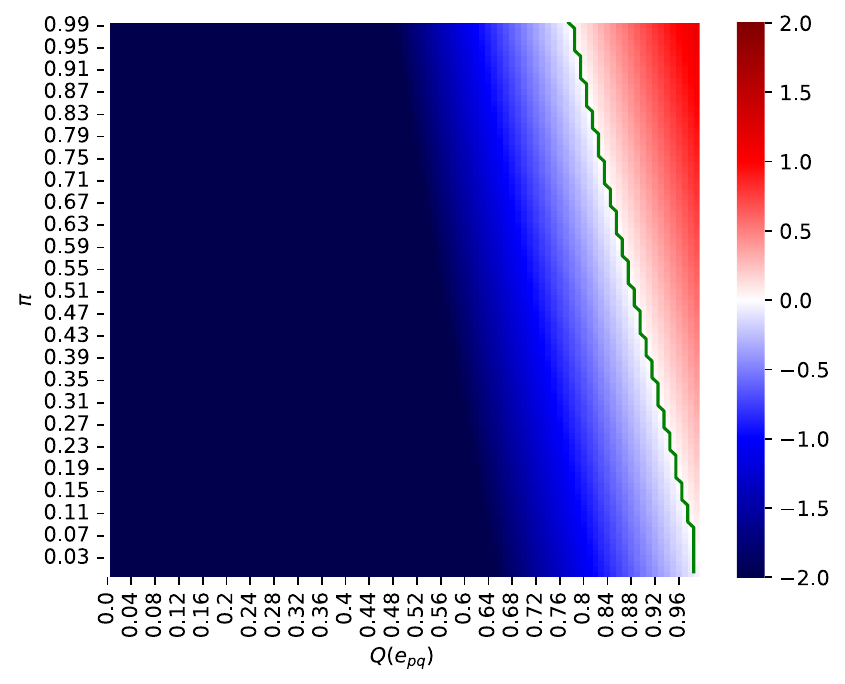}
        \caption{Heatmap visualization of function $\aleph(Q(e_{pq}),\pi)$. The green line is the critical curve of the function, which is less than 0 on the left and greater than 0 on the right. The colors (blue and red) indicate the value of the function.}
        \label{fig:heatmap}
    \end{figure}

Next, we will quantitatively analyze the conditions under which formula (\ref{f:ieq3}) works. We define the function $\aleph(Q(e_{pq}),\pi)=\Gamma Q(e_{pq})-ln(Q(e_{pq}))-ln(Z)-\iota(\rho_1 u_1+\rho_2 u_2)+\pi$. Based on the Person Trajectory Dataset, using Resnet-50 as the visual feature extraction model, we plot the value of $\aleph(Q(e_{pq}),\pi)$ associating with varying $Q(e_{pq})$ and $\pi$ in Fig.\ref{fig:heatmap}. From Fig.\ref{fig:heatmap}, we can see that when $Q(e_{pq})$ is tiny, no matter what value $\pi$ takes, $\Lambda^1(e_{pq})$ will not get any improvement. When $Q(e_{pq})$ gradually increases to a certain value, if $\pi$ exceeds the critical value, then $\Lambda^1(e_{pq})$ is greater than $Q(e_{pq})$. It shows that even if the spatio-temporal model does not have sufficient evidence to confirm that vertex $v_p$ and $v_q$ belong to the same trajectories when they have similar spatio-temporal probability with other nodes, the CRF module tends to think that they belong to the same trajectories. Similarly, even if $Q(e_{pq})$ is large enough, if $\pi$ is small, CRF will judge that the high spatio-temporal probability of current node $p$ and node $q$ is false positive so that $\Lambda^1(e_{pq})$ will be less than $Q(e_{pq})$. The analysis results with more feature extraction models will be reported in the experiment section.


    \begin{figure}[t!]
        \centering
        \includegraphics[width=0.4\textwidth]{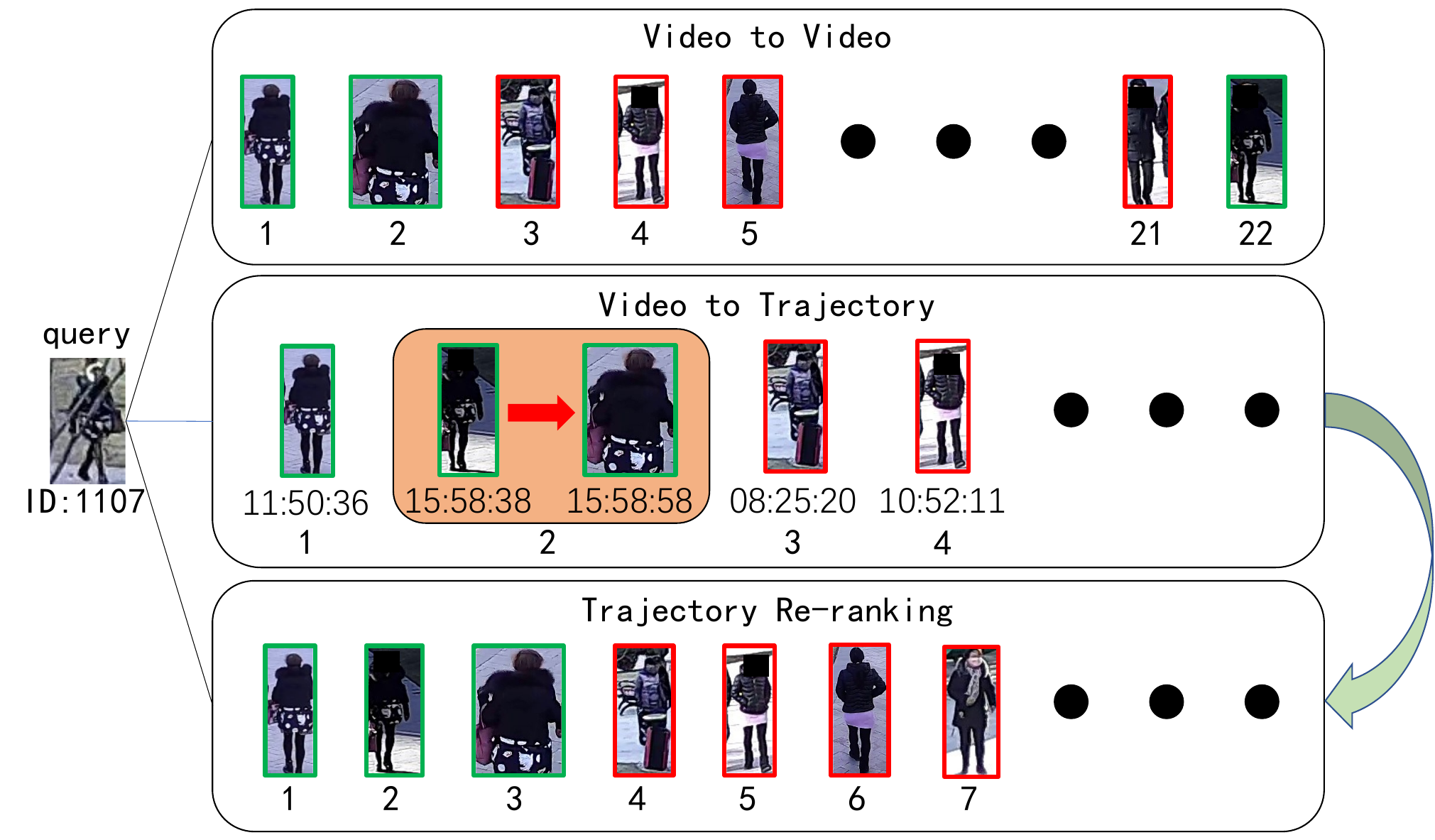}
        \caption{Examples of Video-to-Video retrieval, Video-to-Trajectory retrieval, and Trajectory Reranking. Each image represents the first frame of a tracklet. An image with a green border indicates that its identity is the same as the query, while an image with a red border means its identity differs from the query.}
        \label{fig:rerankingexample}
    \end{figure}

    \subsection{Trajectory Re-ranking for Person Retrieval}
    \label{rerank}
    In the previous stage, we generated multiple possible trajectories. We define $S=\{S_1, S_2,..., S_{N_S}\}$ to represent the generated trajectory gallery, where $S_i$ represents the $i$-th trajectory in $S$. Each trajectory is composed of multiple tracklets, and its feature vector $f_{S_i}$ can be expressed as the average feature vector of all tracklets in $S_i$. Then we calculate the cosine distance between $f_{query}$ and each feature $f_{S_i}$ in $S$ and sort them from smaller to larger to get trajectory retrieval results. In Fig. \ref{fig:rerankingexample}, we use Video-To-Video to indicate that the query and gallery are composed of video clips and Video-To-Trajectory to suggest that the query is video and gallery is composed of trajectories. From Fig. \ref{fig:rerankingexample}, 

when we use the trajectory as the object to be retrieved, we can improve the retrieval accuracy compared with that using only visual features.However, we cannot directly compare the performance between Video-to-Video and Video-to-Trajectory quantitatively because they have different forms of retrieval results. To solve this problem, we use a flattening operation to transform the form of Video-to-Trajectory retrieval results into the form of Video-to-Video retrieval results. Firstly, we assume that $z_1,z_2,z_3,...,z_n$ result from person trajectory retrieval. Among them, the order of $z_i$ is sorted by visual similarity, with high similarity in the front and low similarity in the back. At the same time, all tracklets in each trajectory are sorted by their similarity with the query. Then we connect all the trajectories to get a new sequence and traverse the new sequence from front to back. If the current node has not appeared before, it will be retained. Otherwise, it will be deleted. The results of trajectory re-ranking are illustrated in Fig. \ref{fig:rerankingexample}. It can be seen that the search results of Video-to-Trajectory are successfully transformed into the form of Video-to-Video retrieval results.

    \subsection{Query Spatio-Temporal Re-ranking}
    \label{sec:qstr}
    Query spatio-temporal information is also precious information. The previous works\cite{huang2016camera,wang2019spatial,xie2020progressive} use query spatio-temporal information to re-rank the retrieval results to optimize the performance. In our framework, the query spatio-temporal re-ranking method can also be employed to improve the performance of Video-to-Trajectory retrieval. Given a query, we can get a retrieval result sequence $z_1, z_2,..., z_{N_S}$, just as drawn in Video-to-Trajectory of Fig. \ref{fig:rerankingexample}. We use $d_{S_{z_i}}$ to represent the cosine distance between the $f_{query}$ and $S_{z_i}$. Then we use the following formula to calculate the spatio-temporal probability $\eta_{z_i}$ between the query and $S_{z_i}$:
    \begin{equation}
        \eta_{S_{z_i}}=max\{Q(query, o^{\alpha_k}_j)|o^{\alpha_k}_j \in S_{z_i}\}
    \end{equation}
    where $Q(query,o^{\alpha_k}_j)$ calculates the spatio-temporal probability between the query and tracklet $o^{\alpha_k}_j$ using the MLP spatio-temporal model, and $max$ represents that we take the highest spatio-temporal probability as the spatio-temporal probability between the query and the trajectory $S_{z_i}$. Then we calculate the new similarity $sim^*_{S_{z_i}}$ between the query and $S_{z_i}$ according to the following formula:
    \begin{equation}
        sim^*_{S_{z_i}}=\frac{1}{1+\lambda_1 e^{-h_1 d_{S_{z_i}}}}\times \frac{1}{1+\lambda_2 e^{-h_2 \eta_{S_{z_i}}}}
    \end{equation}
    where $\lambda_1,\lambda_2,h_1,h_2$ are hyper-parameters. Finally, we re-ranking $S$ according to the new similarity $sim^*_{S_{z_i}}$ to obtain a new retrieval result $z^*_1,z^*_2,...,z^*_{N_S}$.

    \section{Experiment}
    \label{sec:experiment}
    The experiments focus on investigating six aspects: i) Comparison of performance between our method with the traditional person retrieval methods; ii) Ablation study for trajectory generation; iii) Effectiveness of trajectory re-ranking; iv) Effectiveness of strategy of applying the spatio-temporal model; v) Spatio-temporal robustness of the proposed method; vi) The generalization ability of the proposed method.

    \subsection{Evaluation Criteria}
    \label{EC}
    Besides evaluating the person retrieval or re-identification, the purpose of PTD is used to evaluate the spatio-temporal modeling and trajectory generation. For the spatio-temporal model evaluation, the Receiver Operating Characteristic Curve (ROC) and the Area Under Curve (AUC) are used. For the person retrieval evaluation, we use three indicators to evaluate: Cumulative Matching Characteristics (CMC), mean Average Precision (mAP), and Trajectory Rank Score (TRS). For CMC and mAP, we use the standard settings \cite{xie2020progressive,huang2016camera,wang2019spatial,MGN,PCB,HACNN} on them to conduct experiments. Trajectory Rank Score (TRS) is proposed in this paper to measure the performance of the trajectory generation algorithm in person retrieval task. TRS is defined as follows.

    Suppose that the input query feature vector is $f_{query}$ and the retrieval result can be represented by $S=\{S_{z_1},S_{z_2},...S_{z_{N_S}}\}$. We define the ground-truth trajectories as $S^*=\{S^*_1, S^*_2,..., S_{N_{S^*}}\}$ where $S^*_i$ represents a ground-truth trajectory with the identity as same as the query and define the number of occurrences of $o^{\alpha_k}_i$ in $S$ as $n_{o^{\alpha_k}_i}$. Then the weight of $o^{\alpha_k}_i$ can be calculated by following formula:
  
    \begin{equation}
    \centering
           g(o^{\alpha_k}_i)=\frac{1}{n_{o^{\alpha_k}_i}}
    \end{equation}
    Then we define the intersection score of the ground truth $S_l^*$ and the prediction $S_{z_j}$ as follows:
    \begin{equation}
       \lvert S^*_l \bigcap S_{z_j} \lvert=\sum_{o^{\alpha_k}_i \in S_{z_j}} g(o^{\alpha_k}_i)\mathbb{I}_{o^{\alpha_k}_i}
    \end{equation}
    where $\mathbb{I}_{o^{\alpha_k}_i}$
     indicates that wether the identity of $o^{\alpha_k}_i$ is the same as the identity of $S^*_l$. Then we define union score:
    \begin{equation}
       \lvert S_l^* \bigcap S_{z_j} \lvert=\sum_{o^{\alpha_k}_i \in S_{z_j}}\mathbb{I}_{o^{\alpha_k}_i}
    \end{equation}
    Finally, we define an intersection ratio operation $\sigma $ to calculate the similarity between the two trajectories:
    \begin{equation}
        \sigma(S^*_l,S_{z_j})=\frac{ \lvert S^*_l \bigcap S_{z_j} \lvert}{ \lvert S^*_l \bigcup S_{z_j} \lvert}
    \end{equation}
    \begin{equation}
        TRS=\sum_{l=1}^{N_{S^*}}\sum_{j=1}^{N_S}\frac{1}{j^2}\sigma(S_{S^*_l},S_{z_j})
    \end{equation}
    Obviously, a higher TRS indicates a better method.

    \subsection{Experiment Setting}
    \label{Experiment_setting}
    Our experiment mainly uses our dataset Person Trajectory Dataset. We choose Resnet-50 as our backbone, from which we removed the last two layers to get the feature extraction model. First of all, we conduct supervised training on the training set of PTD, with the loss function of triplet loss, a learning rate of 0.0002, and a decay rate of 0.0005. The Adaptive Moment Estimation (ADAM) method is used to optimize the model. We randomly select 64 identities in each mini-batch and four images for each identity. A total of 256 images are sent to the network for training, and we train 50 epochs in total, with the learning rate decreased to one-tenth in 20 and 40 periods. To train the spatio-temporal model, we use a multi-layer perceptron composed of a three-layer neural network to train. The input layer inputs the time difference and distance. The middle layer contains 100 leaf nodes. The output layer produces the probability of the data belonging to a particular trajectory using a specific set of inputs.

    \subsection{Person Retrieval Performance}
    \label{advtanges}
\begin{figure}[t!]
        \centering
        \includegraphics[width=0.5\textwidth]{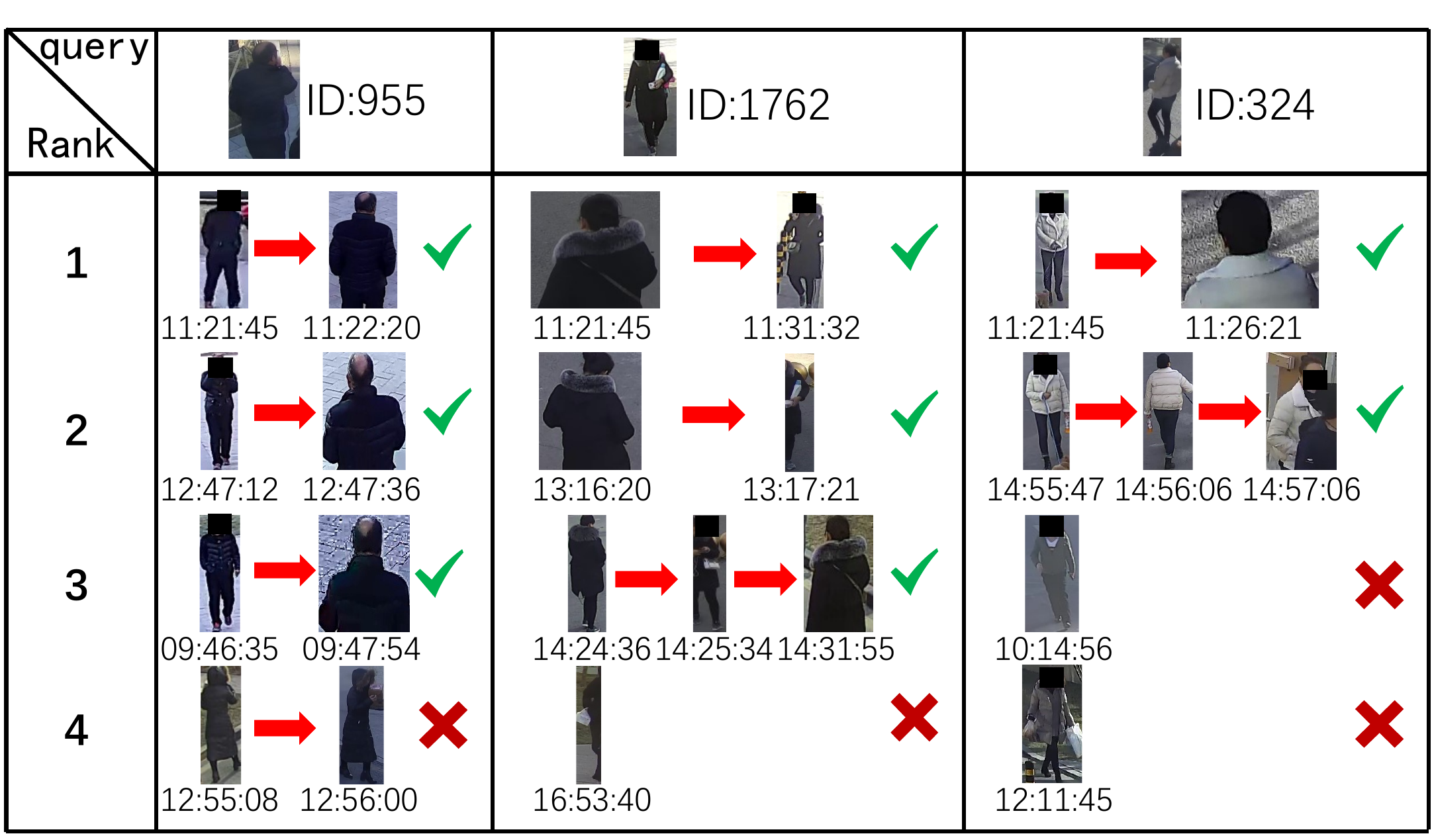}
        \caption{Examples of pedestrian trajectory retrieval. A green tick indicates that the trajectory identity is the same as the query, and a red cross indicates that the trajectory identity differs from the query. Each photo represents a tracklet and is the first picture of the corresponding tracklet.}
        \label{fig:moreexample}
    \end{figure}

\begin{figure}[!t]
    \centering
    \includegraphics[]{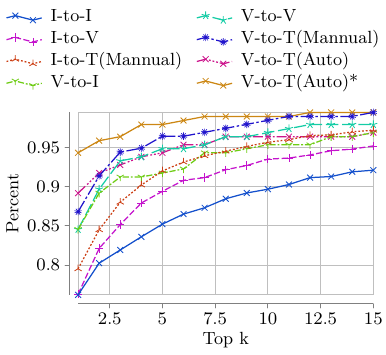}
    \caption{CMC curves on test set of PTD.\label{CMCOFPTD}}
    \end{figure}

    In this section, we experiment with person retrieval strategies using different information. This experiment will verify that using more additional information, such as association under a single-camera and cross-camera trajectory information, can significantly improve the performance of pedestrian retrieval. According to the different information used in the retrieval process, we divide the person retrieval into three types, image-based retrieval, video tracklet-based person retrieval, and trajectory-based person retrieval. When query and gallery are both images, it is called image-based method. When one of the query or the gallery is a video, we call it the video-based method. When there are cross-camera trajectories in the gallery, we call it the trajectory-based method. The experimental results are shown in Table \ref{PSE}. In Table \ref{PSE}, the ``Manual'' represents that we use the manually labeled trajectories for trajectory retrieval and the ``Auto'' represents that we use trajectories automatically obtained by our trajectory generation algorithm for trajectory retrieval. The ``*" indicates that the method utilizes the query spatio-temporal re-ranking described in Section \ref{sec:qstr}. For the image-based method, we apply the most commonly used configuration in the Re-ID task \cite{MGN,PCB} to the experiment. For the video-based method, we use the image-based model to extract features from multiple frames of the same person in a video and take the average as the feature of the video. Finally, we use the same configuration as the image method to test performance. For the trajectory-based method, we first use the manually labeled cross-camera trajectory, extract the average feature vector of all the images in the trajectory as the trajectory feature, then compare with the query feature and sort by their similarity. From Table \ref{PSE}, we can see that V-to-T(Manual) method outperforms the V-to-V method by 4.3\%, 2.3\%, 1.6\%, 1.6\% in mAP, Rank-1, Rank-5, and Rank-10, respectively. From Fig. \ref{CMCOFPTD}, the best performing method is the Video-to-Trajectory method. These experiments show that person retrieval using trajectory information is better than pedestrian retrieval using only video and image information. We also experiment by using an automatically generated trajectory. Without query spatio-temporal information, we can see that our V-to-T(Auto) method is only 0.8\% less than the manual-annotated trajectory on mAP but 3.4\% higher on Rank-1. Under the condition of knowing the spatio-temporal information of the query, it can be seen that our V-to-T(Auto)* method is 2.5\% higher than the one by using manual-annotated trajectory on mAP, 7.5\% higher on Rank-1. These show that our method with an automatically generated trajectory is close to the performance obtained by manual annotation. In Fig. \ref{fig:moreexample}, we give more examples of trajectory retrieval, from which we can see that our method successfully retrieves the pedestrian trajectories of the targets.

    \begin{table}[!t]

    \caption{RESULTS OF PERSON RETRIEVAL USING DIFFERENT MATCHING PATTERNS. HERE ``I", ``V", AND ``T" INDICATE THE ``IMAGE", ``VIDEO", AND ``TRAJECTORY", RESPECTIVELY.}
    \label{PSE}
    \centering
    \begin{tabular}{c|cccc}
    \hline
    Matching Pattern & mAP  & Rank-1 & Rank-5 & Rank-10 \\
    \hline
    I-to-I      & 64.7 & 76.2   & 85.2   & 89.7    \\
    V-to-V      & 85.6 & 84.5   & 94.8   & 96.9    \\
    V-to-I      & 74.1 & 84.5  & 91.8   & 95.4    \\
    I-to-V      & 78.2 & 76.2   & 89.4   & 93.5    \\
    I-to-T(Manual) & 83.7 & 79.5   & 92.0   & 95.7    \\
    V-to-T(Manual) & 89.9 & 86.8 & 96.4 & 98.5 \\
    V-to-T(Auto)   & 89.1 & 90.2 & 94.8 & 95.9  \\
    V-to-T(Auto)* & \textbf{92.4} & \textbf{94.3} & \textbf{97.9} & \textbf{99.0}  \\
    \hline
    \end{tabular}
    \end{table}
    \subsection{Ablation Experiment for Trajectory Generation}
    \label{exp:PTGE}
    In this section, we investigate the different modules in trajectory generation. For the optimization stage of the trajectory generation module, we performed ablation experiments on the CRF module. For the trajectory optimization module, in addition to the RNMF module used in this paper, we compare its alternative Graph Search (GS) algorithm. The GS is a greedy algorithm. Considering the output result of the candidate set generation module graph $G(V,E)$, we first define a trajectory's start and end points. The starting point should have zero in-degree, while the ending should have zero out-degree. By traversing the nodes in $G$, we can get a set $K$ of all vertices with in-degree 0, and we put them to set $V_{start}$. Using the same method, the nodes with out-degree 0 are put to form the terminal set $V_{end}$. Then we use the Depth-First-Search algorithm (DFS) to get the final trajectories. Finally, we use the query feature to compare these trajectory features again to get their similarity. The final sorted result is the final result of person trajectory retrieval.

    In terms of experimental parameter setting, for HC, we set the hyper-parameter threshold to 0.07. When the conditional random field module is used in the experiment, we set the number of iterations $P$ as 1, $\rho_1$ as 1.69, $\rho_2$ as 1.5, $u_1$ as 0, $u_2$ as 0, and $\iota$ as 0.82, $\kappa$ as 0.9. When we use the RNMF method to solve the trajectory, we set the number of iterations of the parameter to 200, $\varpi$ to 0.5, and $\tau$ to 4. 
    

    In terms of quantitative comparison criteria, we select four indicators to compare the performance of different methods: TRS, mAP, Rank, and ANOPN. The ANOPN represents the Average Number of Occurrences Per Node. It represents the sum of the nodes of all generated trajectories divided by the number of nodes in the gallery. The results are shown in Table \ref{table:ablation1} and Table \ref{table:ablation2}. In Table \ref{table:ablation1}, we compare the results of the conditional random field module with different spatio-temporal models. When using MLP spatio-temporal model, the conditional random field method can achieve the best results. The effect of the AdaBoost and DecisionTree spatio-temporal model is not obvious after adding the conditional random field, and the effect of MLP is obviously improved after adding the conditional random field. In Table \ref{table:ablation2}, we compared the cooperation degree between conditional random field and RNMF under the condition of using the MLP spatio-temporal model. The baseline of our experiment is the result obtained in the V-to-V matching pattern without any post-processing. M0 adds a clustering method on the baseline, does not use any spatio-temporal model, and directly takes the clustering results as the trajectory. Compared with the baseline, M0 increased by 0.7\% and 1.6\% on mAP and Rank-1, respectively. This shows that using the results of simple clustering as trajectories is effective. Compared with M0, M1 has better results after adding the graph search method. This indicates that there are incorrect clustering results in the clustering results, and incorrect clustering results can be effectively excluded by adding the graph search method with the spatio-temporal model. However, this also introduces an additional defect: nodes may appear in multiple trajectories, and ANOPN has increased from 1.0 to 1.3. Therefore, we replace GS with RNMF to obtain M2. Compared with M0, this method also improves when ANOPN is kept at 1, and TRS is increased to 0.80, but mAP and Rank-1 are weaker than M1. Based on M1 and M2, we add conditional random field modules to obtain M3 and M4, and the performance has been further improved. Compared with M1, M3 increased TRS by 0.02, mAP by 0.2\%, Rank-1 by 0.6\%, and ANOPN decreased by 0.1. Comparing M4 with M2, TRS increased by 0.02, mAP increased by 1.8\%, Rank-1 increased by 2.6\%, and ANOPN remained unchanged. These two comparisons show that our conditional random field method can effectively mine the spatio-temporal information in the gallery, reduce the clustering errors and improve the retrieval performance. Finally, comparing M3 and M4, there is little difference between the two methods in TRS, mAP, and rank-1. Meanwhile, ANOPN decreased by 0.3, which shows that RNMF can achieve the best results without repeated nodes in the solving stage of the trajectory generation process.

    \begin{table}[t!]
        \caption{THE ABLATION EXPERIMENT OF CRF AND SPATIO-TEMPORAL MODEL.\label{table:ablation1}}
        \centering
        \begin{tabular}{l|l|ccccc}
        \hline
             \makecell[c]{Spatio-temporal\\modeling} & CRF &  TRS$\uparrow$  &mAP$\uparrow$ &Rank-1$\uparrow$\\ \hline
             None         &  -          & 0.77 & 84.0 & 85.1   \\
             Adaboost     &  -          & 0.70 & 83.6 & 84.0   \\
             DecisionTree &  -          & 0.70 & 83.5 & 84.0   \\
             MLP          &  -          & 0.80 & 86.0 & 86.6   \\
             Adaboost     &  \checkmark & 0.75 & 85.6 & 85.1   \\
             DecisionTree &  \checkmark & 0.79 & 84.8 & 85.1   \\
             \textbf{MLP} &  \checkmark & \textbf{0.82} & \textbf{87.8} & \textbf{89.2}   \\
             \hline
        \end{tabular}
    \end{table}

    \begin{table}[]
        \caption{THE ABLATION EXPERIMENT OF CRF AND NRMF.\label{table:ablation2}}
        \centering
        \resizebox{.95\columnwidth}{!}{
        \begin{tabular}{c|c|cc|cccc}
        \hline
             Model & CRF         & RNMF          & GS         & TRS$\uparrow$  &mAP$\uparrow$   &Rank-1$\uparrow$  &ANOPN$\downarrow$\\ \hline
           Baseline& -           & -             & -          & -             & 85.6 & 84.5 & - \\
             M0    & -           & -             & -          & 0.77           & 86.3 & 86.1     & \textbf{1.0}\\
             M1    & -           & -             & \checkmark & 0.77           & 87.1 & 88.1     & 1.4    \\
             M2    &  -          & \checkmark    &  -         & 0.80             & 86.0 & 86.6     & \textbf{1.0}  \\
             M3    &  \checkmark & -             &\checkmark  & 0.79           & 87.3 & 88.7     & 1.3     \\
             \textbf{M4}    &  \checkmark & \checkmark    &-           & \textbf{0.82}           & \textbf{87.8} & \textbf{89.2}    & \textbf{1.0}  \\
             \hline
        \end{tabular}
        }
    \end{table}

    \subsection{Comparison of Feature Re-ranking Methods}
    \label{ETIF}
    In this experiment, we investigate the effectiveness of feature re-ranking and trajectory re-ranking. As a comparison, we report the results of different matching patterns, Video-to-Video, and Video-to-Trajectory, respectively, together with different re-ranking methods. Note that the trajectory re-ranking is necessary for the trajectory-based retrieval. The experimental results are shown in Table \ref{tab:ERUFRC}. As shown, two kinds of re-ranking operations both improve the retrieval results. The model with two kinds of re-rankings simultaneously performs better than the one with only one kind of re-ranking.

    \begin{table}[t!]
    \centering
    \caption{PERSON RETRIEVAL RESULTS UNDER DIFFERENT RE-RANKING METHODS. HERE ``V" AND ``T" INDICATE VIDEO AND TRAJECTORY, RESPECTIVELY.\label{tab:ERUFRC}}
    \begin{tabular}{c|c|c|ccc}
    \hline
    \makecell[c]{Matching\\pattern}  & \makecell[c]{Feature\\re-ranking} & \makecell[c]{Trajectory\\re-ranking} & mAP$\uparrow$ & Rank-1$\uparrow$  & TRS$\uparrow$\\
    \hline
    V-to-V &    -            & -                  &  85.6            & 84.5   &   -\\
    V-to-V &   \checkmark    &          -         &  88.9            &  88.7  & -    \\
    V-to-T &    -            &    \checkmark      &  87.8            &   89.2 & 0.82    \\
    V-to-T &   \checkmark    &     \checkmark     &   \textbf{89.1}  &   \textbf{90.2} &\textbf{0.83}   \\
    \hline
    \end{tabular}
    \end{table}
    \subsection{Strategies of Using Spatio-Temporal Information}
    \label{CWOSPTS}
    This section compares our strategy using spatio-temporal information with the other strategies. We summarize that strategies for applying spatio-temporal information in pedestrian retrieval are mainly through the Bayesian method, which assumes that the node similarity between the query and the gallery meets the following relationship:
    \begin{equation}
        \label{f:sss}
        Score^*=Score_{V} \times Score_{ST}
    \end{equation}
    where $Score^*$ indicates the comprehensive score, $Score_V$ is the visual similarity between the query and the gallery, and $Score_{ST}$ is the spatio-temporal probability. The Bayesian strategy includes two steps. The first step is calculating the query and gallery's visual similarity and spatio-temporal probability. The second step is calculating the total score through formula (\ref{f:sss}) and sorting it according to the $Score^*$ to get the search results. Therefore, the disadvantage of using the Bayesian strategy is that we must know the spatio-temporal information of the query in advance. We reproduce three spatio-temporal re-ranking methods that use the Bayesian strategy for comparison: i) Huang et al.\cite{huang2016camera} used a priori Weibull distribution to get the spatio-temporal model of the camera network. ii) Wang et al.\cite{wang2019spatial} used a statistical histogram model to get the spatiotemporal model of the camera network. Moreover, Laplace smoothing is added to the Bayesian strategy in this method. ii) Xie et al.\cite{xie2020progressive} proposed to use a priori Gaussian distribution to fit the temporal and spatial distribution of the camera network. We apply their Bayesian strategy and corresponding spatiotemporal model to the PTD dataset proposed in this paper, and the experimental results are shown in Table \ref{CEO}. It can be seen that in addition to Wang's method, the performance of other methods based on the priori probability model has decreased dramatically. We summarize the following reasons: i) Our dataset covers larger cameras and a long time. The simple use of the priori probability model cannot fit our dataset's temporal and spatial distribution; ii) Wang's method effectively adds Laplace smoothing to the Bayesian strategy. We replaced the spatio-temporal model in Wang's method with the MLP model, which also achieved good results. This shows that our spatio-temporal model is also effective for the Bayesian strategy, and our MLP spatiotemporal model uses only a few camera pairs to fit the data. In contrast, the histogram model needs to fit the data with all camera pairs. Next, we put the effective histogram model in Bayesian strategy into our strategy and experiment with conditional random field and RNMF method. It can be seen that the histogram spatio-temporal model has only achieved a small improvement in our strategy. This shows that the histogram model is weaker than the MLP model in mining spatio-temporal information in the camera network. Finally, we combine Wang's Bayesian strategy with our strategy, and the results are better than any single strategy. This shows that our strategy and the existing Bayesian strategy are not mutually exclusive but promote each other.

\begin{figure}[t!]
        \centering
         \subfloat[]{\includegraphics[width=0.25\textwidth]{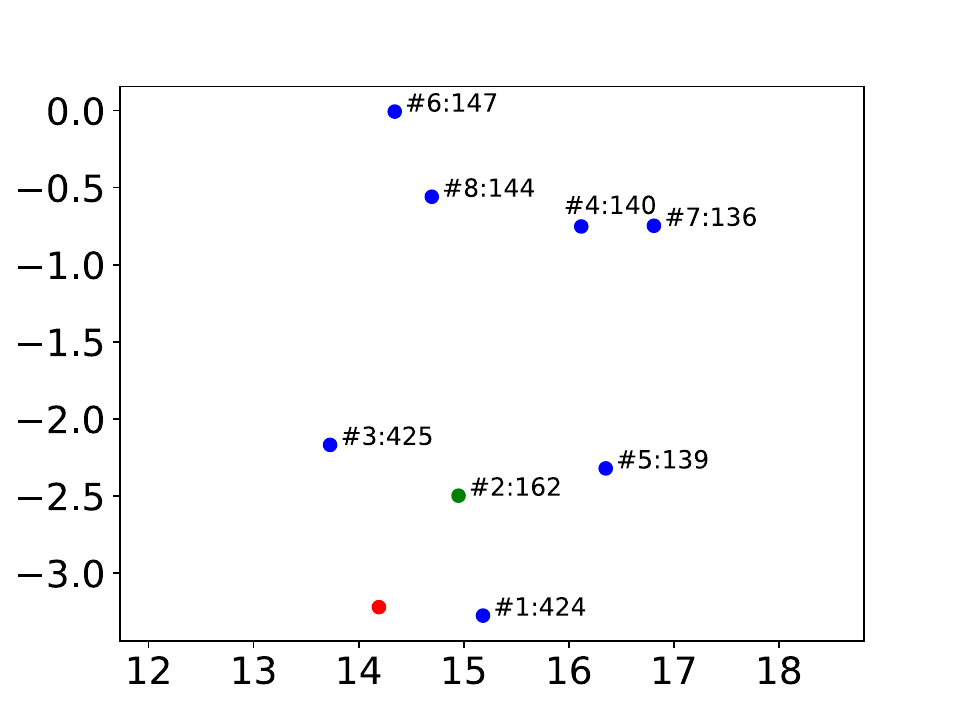}\label{visa}}
        \subfloat[]{\includegraphics[width=0.25\textwidth]{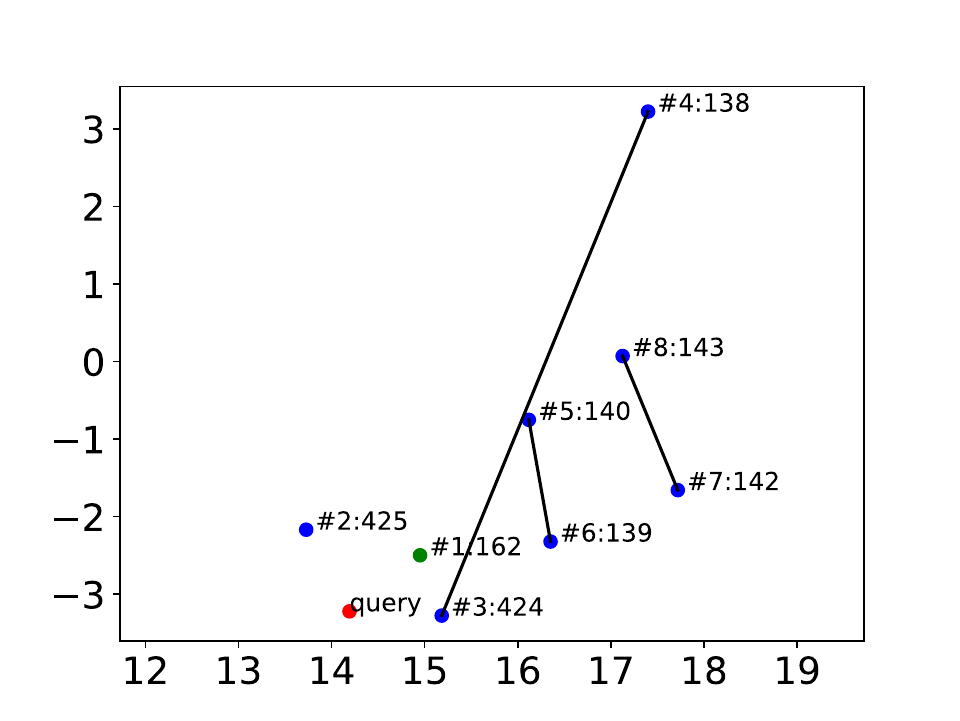}\label{visb}}
        \caption{T-SNE visualization of our strategy. Red circles represent the query, green circles represent positive samples, and blue circles represent negative samples. The number after ``\#'' indicates its order in this sorting, and the number after ``:'' indicates the corresponding tracklet ID. (a)The original sorting result. (b) The sorting results after using our strategy. Nodes connected by straight lines indicate that they belong to a trajectory.}
        \label{fig:vis}
    \end{figure}

    Next, we visualize our strategy. The visualization results are shown in Fig. \ref{fig:vis}. We select an example from the test set to visualize the effectiveness of our method. It shows the results of the visualization of the query and its most similar top 10 in the gallery using T-SNE. It can be seen that the nearest sample to query is the wrong sample, ID 424, and the correct sample should be ID 162. When we use the strategy proposed, we will establish the pedestrian trajectory in the gallery. It can be seen that the nearest sample ID 424 belongs to the trajectory with sample ID 138, and their center is farther away from the query than sample ID 162, so sample ID 162 becomes the top 1.

    \begin{table*}[t!]
        \centering

        \caption{COMPARISON WITH OTHER SPATIO-TEMPORAL STRATEGIES}
        \begin{tabular}{c|c|c|c|c|ccccc}
        \hline
             \makecell{Matching \\Pattern}&Method      & \makecell{Spatio-Temporal \\Strategy}& \makecell{Spatio-Temporal\\Model }&\makecell{Query Spatio\\-Temporal Information} &mAP$\uparrow$ &Rank-1$\uparrow$&TRS$\uparrow$ \\ \hline

            \multirow{4}{*}{V-to-V} & Baseline & - & -  & No& 85.6 &84.5 &-\\
            & Huang et al.\cite{huang2016camera} & Bayesian & Weibull distribution & Yes & 45.2 & 25.8  &-\\
            & Wang  et al.\cite{wang2019spatial} & Bayesian &  Histogram model & Yes& 92.2 &  94.3 &- \\
                    & Wang  et al.\cite{wang2019spatial} & Bayesian & MLP & Yes& 91.5 &  \textbf{94.8} &- \\
             &Xie et al.\cite{xie2020progressive} & Bayesian & Gaussian distribution & Yes  & 68.1&78.4 &-\\ \hline
            \multirow{4}{*}{V-to-T} & Ours & Clustering and Scattering & MLP & No & 87.8 & 89.2 & 0.82\\
             & Ours & Clustering and Scattering & Histogram model & No & 85.8 & 87.1 & 0.77\\
              & \makecell{Wang  et al.\cite{wang2019spatial}\\+Ours}&\makecell{Bayesian\\+Clustering and Scattering} & \makecell{MLP} & Yes& \textbf{92.4} & 94.3 & \textbf{0.87} \\ \hline
        \end{tabular}
        \label{CEO}
    \end{table*}

    \subsection{Spatio-Temporal Robustness}
\begin{figure}[t!]
        \centering
        \includegraphics[width=0.4\textwidth]{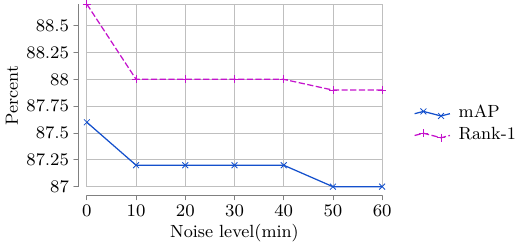}
        \caption{The influence of spatio-temporal noise on the proposed method.}
        \label{fig:noise}
    \end{figure}

    In this section, we study the influence of spatio-temporal noise on the method in this paper. Firstly, we analyze the source of spatio-temporal noise. We conclude that spatio-temporal noise mainly comes from the following ways: i) the walking speed of different pedestrians is varying; ii) There are different paths between a pair of cameras; iii) We cannot judge whether a pedestrian is moving forward or doing something else when she is not in the camera. To study the influence of the above conditions on the method in this paper, we add different levels of random uniform noise from 10 minutes to 60 minutes at 10-minute intervals to the spatio-temporal data of the test set. The experiment results are obtained by averaging ten times in each experiment. The experimental results are shown in Fig. \ref{fig:noise}. It can be seen that after adding 60 minutes of random uniform noise, mAP only decreases by 0.6\% and Rank1 only decreases by 0.8\%. Therefore, our method is robust to spatio-temporal noise in the camera network.

    \begin{figure}[t!]
        \centering
        \includegraphics[width=0.4\textwidth]{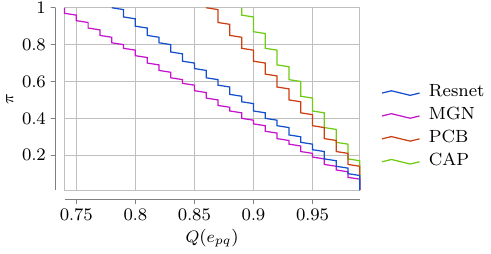}
        \caption{The critical curves of different visual feature extraction models. The upper right corner of the critical curve is the positive gain region and the lower left corner is the negative gain region.}
        \label{fig:total}
    \end{figure}

     \begin{table}
       \centering
       \caption{PERSON RETRIEVAL RESULTS USING DIFFERENT FEATURE EXTRACTION MODELS. HERE ``V" AND ``T" INDICATE VIDEO AND TRAJECTORY, RESPECTIVELY. THE ``FR'' AND ``TR'' INDICATE FEATURE RE-RANKING AND TRAJECTORY RE-RANKING, RESPECTIVELY.}
       \label{tab:GE}
         \begin{tabular}{c|c|c|c|ccccc}
            \hline
         \makecell[c]{Matching\\pattern} & \makecell[c]{FR} & \makecell[c]{TR} & \makecell[c]{Feature\\Model} & mAP$\uparrow$ & Rank-1$\uparrow$  & TRS$\uparrow$\\\hline
         \multirow{4}{*}{I-to-I} & - & - & CAP\cite{CAP}             & 56.1 & 67.7  & -\\
                                 & - & - & MGN\cite{MGN}             & 75.8 & 86.4  & -\\
                                 & - & - & Resnet50\cite{7780459} & 64.7 & 76.2  & -\\
                                 & - & -  & PCB\cite{PCB}            & 59.3 & 73.0 &  - \\ \hline
         \multirow{4}{*}{V-to-V} & - & - & CAP\cite{CAP}             & 81.6 & 79.9  & -\\
                                 & - & - & MGN\cite{MGN}             & 93.0 & 92.8  & -\\
                                 & - & - & Resnet50\cite{7780459} & 85.6 & 84.5  & -\\
                                 & - & -  & PCB\cite{PCB}            & 83.7 & 83.0 &  - \\ \hline

         \multirow{4}{*}{V-to-V} & \checkmark & - & CAP\cite{CAP}    & 81.6 & 79.9  &- \\
                                 & \checkmark & - & MGN\cite{MGN}    & 95.0 & 94.8 &  - \\
                                 & \checkmark & - & Resnet50\cite{7780459} & 88.9 & 88.7 &  -\\
                                 & \checkmark & - & PCB\cite{PCB}    & 84.9 & 84.0  & - \\ \hline

         \multirow{4}{*}{V-to-T} & - &  \checkmark & CAP\cite{CAP} & 83.0 & 81.4 & 0.77 \\
                                 & - & \checkmark & MGN\cite{MGN} & 94.8 & 94.3  & \textbf{0.90} \\
                                 & - & \checkmark & Resnet50\cite{7780459} & 87.8 & 89.2  & 0.82\\
                                 & - & \checkmark & PCB\cite{PCB} & 84.9 & 84.5 &  0.78\\ \hline

         \multirow{4}{*}{V-to-T} & \checkmark & \checkmark & CAP\cite{CAP} & 83.0 & 81.4 & 0.77\\
                                 & \checkmark & \checkmark & MGN\cite{MGN} & \textbf{95.1} & \textbf{95.4}  & \textbf{0.90}\\
                                 & \checkmark & \checkmark & Resnet50\cite{7780459} & 89.1 & 90.2  &0.83\\
                                 & \checkmark & \checkmark & PCB\cite{PCB} & 85.6 & 85.1 & 0.78\\ \hline
         \end{tabular}%
     \end{table}%

    \subsection{Generalization}
    \label{MGE}
    Finally, we are concerned about whether the trajectory generation method and the trajectory re-ranking method proposed in this paper can be applied to other feature extraction models besides Resnet-50. We use more person Re-ID models for feature extraction, including the
     supervised ones, PCB\cite{PCB} and MGN\cite{MGN}, and the unsupervised one, CAP\cite{CAP}. The experimental results are shown in Table \ref{tab:GE}. The proposed trajectory retrieval strategy is effective in all feature extraction models. Compared with the Image-to-Image mode using only visual information, the trajectory retrieval method using our clustering and scattering strategy can greatly improve the retrieval performance. In Video-to-Video mode, without feature re-ranking, the proposed trajectory retrieval strategy can improve the mAP by 1.4\% and Rank-1 by 1.5\% on the CAP model. In the supervised model Resnet50, MGN and PCB, the mAP has increased by 2.2\%, 1.8\%, and 1.2\%, respectively. When using feature re-ranking, MGN achieves the best performance. Compared with pure Video-to-Trajectory, integrating the trajectory re-ranking method to the feature re-ranking method of the MGN model can improve Rank-1 by 0.1 points. When using CAP, Resnet50, PCB model, the mAP has increased by 1.4\%, 3.5\%, 1.9\%, and Rank-1 has increased by 1.5\%, 5.7\%, 2.1\%, respectively.

    We also draw the critical curves of different visual feature extraction models, as shown in Fig. \ref{fig:total}. As can be seen from Fig. \ref{fig:total}, the larger the positive gain area of the model with good visual feature extraction performance, the smaller the positive gain area of the model with poor visual feature extraction. When the visual feature extraction is accurate, there are fewer errors in the generation candidate set. The CRF module will relax the spatio-temporal establishment range of the same trajectory and make the nodes in the candidate set belong to the same trajectory as much as possible. When the visual feature extraction model is poor, there are more errors in the candidate set. The CRF module should use a more strict positive gain region to reduce the errors in the generating trajectory.

    \section{Conclusion}
    \label{sec:Conclusion}

    Towards practical application scenarios, we propose a person retrieval framework based on cross-camera trajectory. In this framework, both visual features and non-visual cues of the camera network are exploited. Specifically, we propose a pedestrian trajectory generation method, which can find the potential scattering patterns in the clustering results of visual feature space and then improve the accuracy of pedestrian retrieval results. In the trajectory generation method, we propose a cross-camera spatio-temporal model that integrates the walking habits of pedestrians and the path distribution between cameras and develop a scheme for trajectory generation based on the spatio-temporal model. We also propose a CRF module to optimize the spatio-temporal probability graph. Then we obtain the specific pedestrian trajectory information by solving the restricted non-negative matrix factorization method. Furthermore, a trajectory re-ranking method is used to enhance the person retrieval accuracy. To verify the proposed method, we build a cross-camera person trajectory dataset from a real scenario.

{\appendix[]
\label{appendix}

 In the appendix, we will introduce how to obtain formula (\ref{f:ieq3}) from the spatio-temporal gain function. Now consider the first iteration, bring formula (\ref{energy}) into formula (\ref{f:gain}), we can get the following formula:
    \begin{equation}
     \label{f:sim}
     \begin{aligned}
    &D(e_{pq})=\frac{e^{ln(ZQ(e_{pq})))}}{Z}\times\\
     &(e^{\iota[\rho_1 (Q(e_{pq})-u_1)+\rho_2(\Upsilon(v_p,v_q)-u_2)]-ln(ZQ(e_{pq})))}-1)
    \end{aligned}
    \end{equation}
    It is easy to see from formula (\ref{f:sim}) that the condition of $D>0$ is equivalent to the following inequality:
    \begin{equation}
     \iota[\rho_1 (Q(e_{pq})\!-\!u_1)\!+\!\rho_2(\Upsilon(v_p,v_q)\!-\!u_2)]\!-\!ln(ZQ(e_{pq})))\!>\!0
     \label{f:ieq}
    \end{equation}
    Bring formula (\ref{f:upsilon}) into formula (\ref{f:ieq}), we can get the following inequality:
    \begin{equation}
    \label{f:ieq2}
     \begin{aligned}
     &\iota(\rho_1\!+\!\frac{\rho_2}{\sqrt{\sum_{s}{Q(e_{ps})^2}}\sqrt{\sum_{s}{Q(e_{qs})^2}}})Q(e_{pq})\!-\!ln(Q(e_{pq}))\!\\
     &\!>\!ln(Z)\!-\!\iota\rho_2\frac{\sum_{s\neq q}{Q(e_{ps})Q(e_{qs})}}{{\sqrt{\sum_{s}{Q(e_{ps})^2}}\sqrt{\sum_{s}{Q(e_{qs})^2}}}}\!+\!\iota(\rho_1 u_1+\rho_2 u_2)
    \end{aligned}
    \end{equation}
    In formula (\ref{f:ieq2}), we take its approximate expression for $\iota(\rho_1\!+\!\frac{\rho_2}{\sqrt{\sum_{s}{Q(e_{ps})^2}}\sqrt{\sum_{s}{Q(e_{qs})^2}}})$ to make it independent of $Q(e_{pq})$:
    \begin{equation}
        \label{f:approx11}
        \begin{aligned}
        \iota(\rho_1\!+\!\frac{\rho_2}{\sqrt{\sum_{s}{Q(e_{ps})^2}}\sqrt{\sum_{s}{Q(e_{qs})^2}}})\approx\\
        \iota(\rho_1\!+\!\frac{\rho_2}{\sqrt{\sum_{s\neq q}{Q(e_{ps})^2}}\sqrt{\sum_{s}{Q(e_{qs})^2}}})=\Gamma
        \end{aligned}
    \end{equation}
    We do the same operation for $\iota\rho_2\frac{\sum_{s\neq q}{Q(e_{ps})Q(e_{qs})}}{{\sqrt{\sum_{s}{Q(e_{ps})^2}}\sqrt{\sum_{s}{Q(e_{qs})^2}}}}$:
    \begin{equation}
        \label{f:approx22}
        \begin{aligned}
        \iota\rho_2\frac{\sum_{s\neq q}{Q(e_{ps})Q(e_{qs})}}{{\sqrt{\sum_{s}{Q(e_{ps})^2}}\sqrt{\sum_{s}{Q(e_{qs})^2}}}}\approx\\
        \iota\rho_2\frac{\sum_{s\neq q}{Q(e_{ps})Q(e_{qs})}}{{\sqrt{\sum_{s\neq q}{Q(e_{ps})^2}}\sqrt{\sum_{s}{Q(e_{qs})^2}}}}=\pi
        \end{aligned}
    \end{equation}
    Bring formula (\ref{f:approx11}) and (\ref{f:approx22}) into (\ref{f:ieq2}), we can get follow formula:
    \begin{equation}
        \Gamma Q(e_{pq})-ln(Q(e_{pq}))>ln(Z)+\iota(\rho_1 u_1+\rho_2 u_2)-\pi
    \end{equation}

 \bibliographystyle{IEEEtran}
 \bibliography{TIP22-ZX.bib}

%








\end{document}